\pdfoutput=1
\documentclass[11pt]{article}

\usepackage[whole]{bxcjkjatype}

\usepackage[preprint]{acl}
\usepackage{times}
\usepackage{latexsym}
\usepackage[T1]{fontenc}
\usepackage[utf8]{inputenc}
\usepackage{microtype}
\usepackage{inconsolata}

\usepackage{comment}
\usepackage{amsmath}
\usepackage{amssymb}
\usepackage{amsfonts}
\usepackage{array}
\usepackage{multirow}
\usepackage{booktabs}
\usepackage{adjustbox}
\usepackage{subcaption}
\usepackage{stmaryrd}
\usepackage{siunitx}
\usepackage{bm}
\usepackage{longtable}
\usepackage{graphicx}
\usepackage{enumitem}
\usepackage{xurl}
\usepackage{fontawesome5}

\newcommand{\bbR}{\mathbb{R}}
\newcommand{\bmxi}{\bm{\xi}}
\newcommand{\bmeta}{\bm{\eta}}

\newcommand{\bmell}{\bm{\ell}}

\DeclareMathOperator*{\Var}{Var}

\DeclareMathOperator*{\bbE}{\mathbb{E}}
\DeclareMathOperator*{\KL}{KL}

\newcommand{\MI}{\textrm{MI}}

\usepackage{xcolor}
\usepackage{pifont}
\newcommand{\cmark}{\textcolor{green!50!black}{\large\ding{51}}}
\newcommand{\xmark}{\textcolor{red!70!black}{\large\ding{55}}}

\newcommand\blfootnote[1]{%
  \begingroup
  \renewcommand\thefootnote{}\footnote{#1}%
  \addtocounter{footnote}{-1}%
  \endgroup
}

\title{Language Model Maps for Prompt-Response Distributions\\via Log-Likelihood Vectors}

\author{
Momose Oyama${}^{\ast\,1,2}$\quad Yusuke Takase${}^{\ast\,1}$\quad Hidetoshi Shimodaira${}^{1,2}$\\
${}^{1}$Kyoto University\quad
${}^{2}$RIKEN\\
\texttt{\{oyama.momose, y.takase\}@sys.i.kyoto-u.ac.jp,}\\
\texttt{shimo@i.kyoto-u.ac.jp}
}
\begin{document}
\maketitle

\begin{abstract}

We propose a method that represents language models by log-likelihood vectors over prompt-response pairs and constructs model maps for comparing their conditional distributions. In this space, squared Euclidean distances between models are approximately proportional to the KL divergence between the corresponding conditional distributions. Experiments on a large collection of publicly available language models show that the maps capture meaningful global structure, including relationships to model attributes and task performance. The representation also captures systematic shifts induced by prompt modifications and their approximate additive compositionality; we use the vectors to predict downstream task scores and leverage their additive structure to approximate the effects of composite prompt operations without directly observing the corresponding log-likelihood vectors. We further introduce PMI vectors to reduce the influence of unconditional distributions; in some cases, PMI-based model maps better reflect training-data-related differences. Overall, the framework supports the analysis and prediction of input-dependent model behavior.
\blfootnote{$^\ast$ Equal contribution.}
\blfootnote{\faGithub~\url{https://github.com/shimo-lab/modelmap}}
\end{abstract}

\section{Introduction}

\begin{figure}[t]
\centering
  \includegraphics[width=\columnwidth]{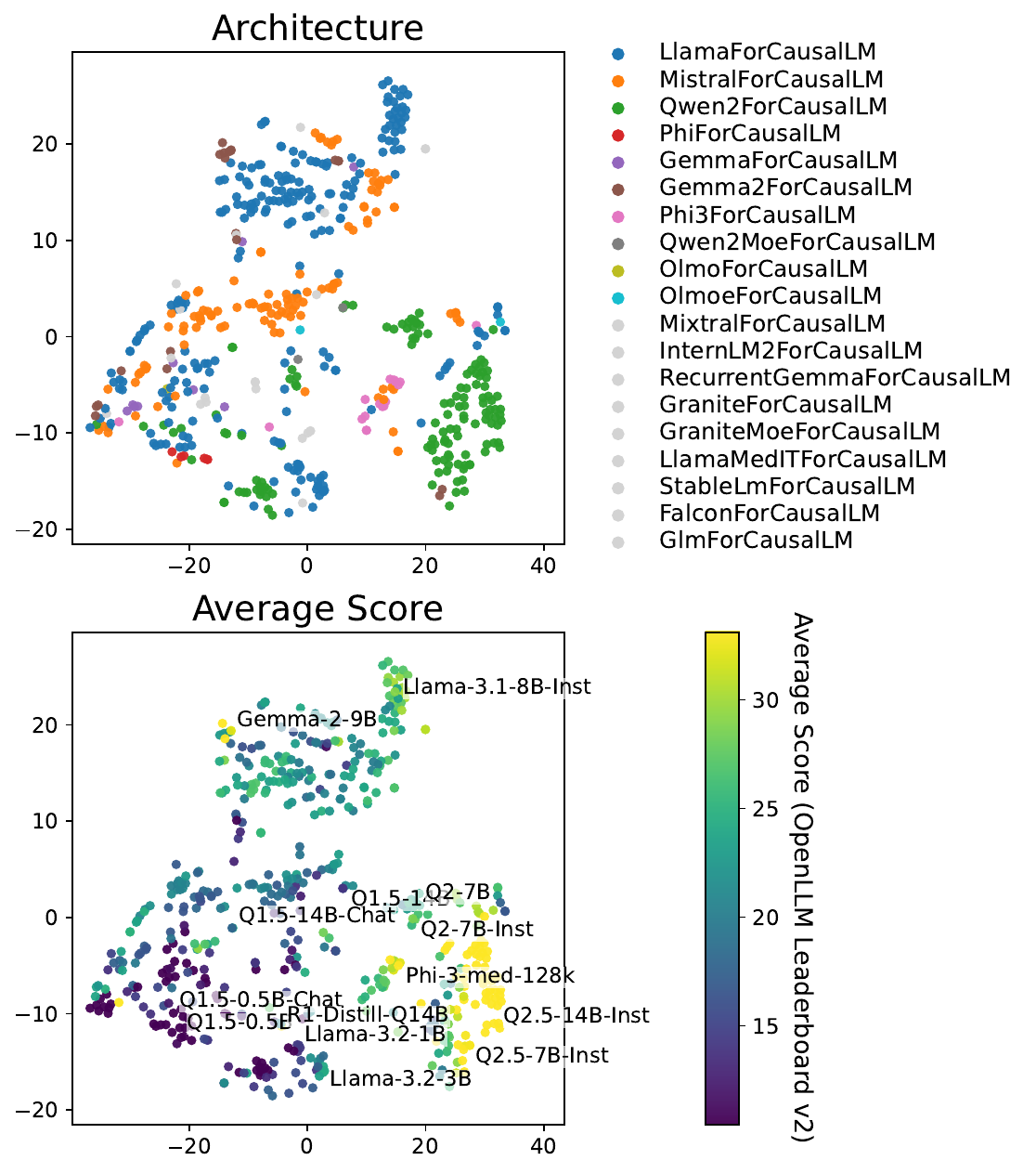}
\caption{Map of $552$ language models. Each model is represented by a conditional log-likelihood vector over prompt-response pairs from the Tulu3 text set and visualized using t-SNE. Top: points are colored by architecture type. Bottom: points are colored by model performance, based on the average score on Open LLM Leaderboard v2. See Section~\ref{sec:conditional-model-map} for details.}
  \label{fig:intro2}
\end{figure}

Large language models (LLMs) are becoming increasingly diverse, creating a growing need for frameworks that systematically compare the behaviors of different models~\cite{zhu2025independence, yax2025phylolm, zhuang2025embedllm, horwitz2025we, jiang2025artificial}. Quantifying distances between models provides a foundation for a wide range of applications, including model selection, performance prediction, and the analysis of distillation and evolution.

A log-likelihood vector~\cite{modelmap2025} is a feature representation that characterizes a language model by its log-likelihoods $\log p(y)$ over a predefined set of texts $y$. In this representation, each language model is embedded as a point in a Euclidean space. The squared Euclidean distance in this space is approximately proportional to the Kullback--Leibler (KL) divergence between models, making distances between models interpretable. Dimensionality reduction of these vectors yields a model map that reflects relationships among probability models and enables model behavior to be analyzed from various perspectives.

However, large language models behave as conditional probability models that define a response distribution $p(y|x)$ conditioned on an input prompt $x$, in contrast to the \textit{unconditional} distribution $p(y)$ used above. Major tasks such as dialogue and question answering are naturally formulated as conditional generation, and model behavior is likewise governed by $p(y|x)$. The term prompt-response distribution refers to a distribution over prompt-response pairs $(x,y)$, such as the joint $p(x,y)$ or the conditional $p(y|x)$. Extending the log-likelihood vector framework to conditional distributions is therefore important for broadening its applicability.

\citet{jiang2025artificial} proposed a method for measuring differences between language models based on distances in the embedding space of response texts generated by the models for a given input. Although this approach directly addresses the practical setting of conditional generation, it can be affected by the randomness of text generation and the variability of embedding representations.

In this work, we propose a method for comparing a large number of language models in the practical setting of conditional generation, without using texts generated by the models themselves. Specifically, for a set of text pairs $D = \{(x_s, y_s)\}_{s=1}^{N}$, we characterize each model by the vector $\{\log p(y_s | x_s)\}_{s=1}^{N}$, which we call a conditional log-likelihood vector. This representation directly evaluates the response distribution conditioned on the input and thus naturally fits the setting of conditional generation. We further show that the squared distance between these vectors approximates the KL divergence between conditional distributions averaged over the input distribution.

Furthermore, to remove the contribution of the unconditional distribution, we introduce a representation based on the difference between the conditional and unconditional log-likelihoods, namely $\log p(y | x) - \log p(y)$. This quantity takes the form of pointwise mutual information (PMI) and can be interpreted as a measure of the strength of the relationship between the input and the response. 
As we show later, model maps based on PMI vectors can in some cases reflect training-data-related differences more clearly than conditional model~maps.

\begin{figure*}[!t]
    \centering
    \includegraphics[width=0.85\linewidth]{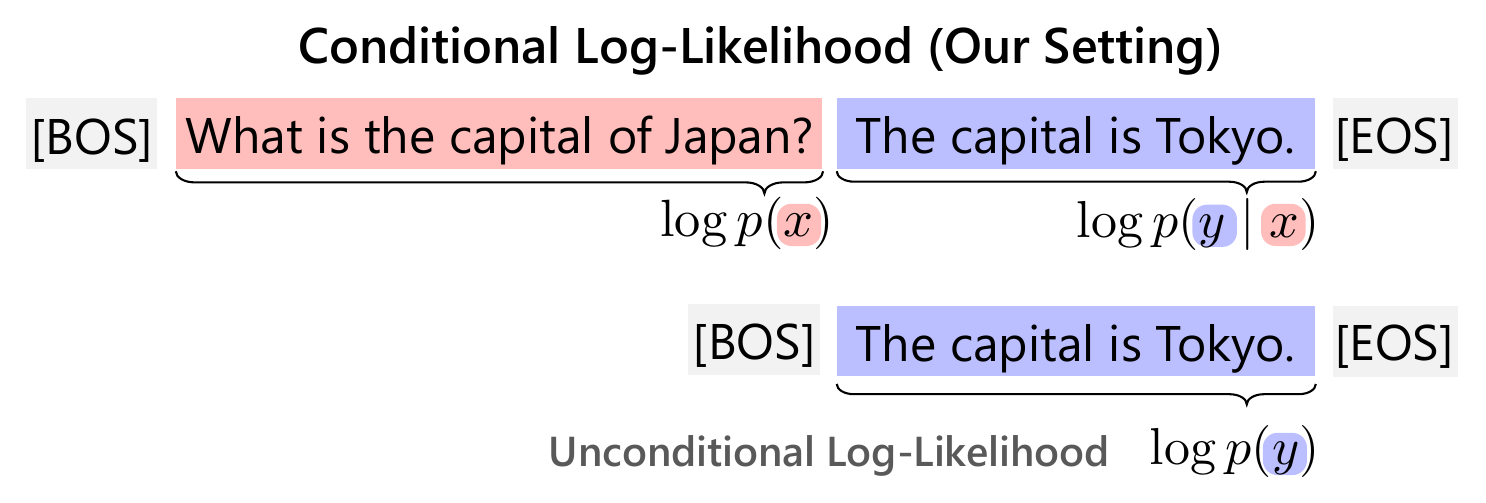}
    \caption{Conditional and unconditional log-likelihoods. Top: a prompt $x$ (red) and a response $y$ (blue) are given to a language model as a single sequence, and the log-likelihood of the sequence decomposes into $\log p(x)$ over the prompt tokens and $\log p(y|x)$ over the response tokens; this paper uses $\log p(y|x)$. Bottom: given the response alone, the model yields the unconditional log-likelihood $\log p(y)$, which is what \citet{modelmap2025} used. Both settings evaluate the same response $y$ and differ only in whether the prompt is given.}
    \label{fig:oursetting}
\end{figure*}

\begin{table*}[!t]
\centering
\begin{adjustbox}{width=0.85\linewidth}
\begin{tabular}{lcl}
\toprule
Method & Prompt & Formulation \\
\midrule
Unconditional Log-Likelihood Vector~\cite{modelmap2025} & \xmark & $\log p(y)$ \\
Conditional Log-Likelihood Vector (Ours) & \cmark & $\log p(y|x)$ \\
PMI Vector (Ours) & \cmark & $\log p(y|x) - \log p(y)$ \\
\bottomrule
\end{tabular}
 \end{adjustbox}
\caption{The three vector representations compared in this paper. The Prompt column indicates whether each component takes the input prompt $x$ into account.}
\label{tab:methods}
\end{table*}

The contributions of this paper are as follows.
\begin{enumerate}
    \item We propose a framework for model maps based on conditional log-likelihood vectors over prompt-response text pairs, and show that distances in this space approximate the KL divergence between conditional distributions averaged over the input distribution.
    \item We show that the proposed shared space captures changes in model position induced by prompt modifications (Prompt Shift) and that these shifts exhibit approximate additive compositionality across composite prompt operations.
    \item Leveraging this geometric regularity, we demonstrate that conditional log-likelihood vectors can be used to predict downstream task performance, and that the effect of an \emph{unobserved} composite prompt can be approximated by combining the shift vectors of its components.
\end{enumerate}

The rest of this paper is organized as follows. Section~\ref{sec:method} introduces our framework for mapping conditional distributions. Section~\ref{sec:conditional-model-map} presents the constructed model maps and analyzes prompt shifts. Sections~\ref{sec:score-prediction} and~\ref{sec:empirical-validation} provide downstream task predictions and empirical validations, respectively. Detailed discussions of related work and experimental settings are deferred to Appendices~\ref{sec:related} and \ref{sec:experimental-settings}.

\section{Our Method: Mapping Conditional Probability Distributions}
\label{sec:method}

In this section, we introduce a method for comparing the conditional probability distributions of autoregressive language models based on conditional log-likelihoods. Figure~\ref{fig:oursetting} shows the overall setting considered in this study, and Table~\ref{tab:methods} summarizes the three representations.

\subsection{Autoregressive language models}
\label{sec:autoregressive}

For a language model $p_i$, we consider the conditional probability distribution $p_i(y | x)$, where $x$ denotes a prompt and $y$ denotes a response.
In an autoregressive language model, $p_i(y | x)$ is factorized as
\begin{equation} \label{eq:piyx}
p_i(y|x) = \prod_{t=1}^{|y|} p_i\bigl(z_t|x, z_{<t}\bigr).
\end{equation}
Here, $|y|$ denotes the number of tokens in the response, and we write $y = (z_1,\dots,z_{|y|})$. In addition, $z_{<t}=(z_1,\dots,z_{t-1})$ denotes the sequence of tokens preceding position $t$.
In the usual use of a language model, the response $y$ is generated by sampling. In contrast, our main use in this paper is to fix the response $y$ in advance and compute its conditional probability using (\ref{eq:piyx}).

\subsection{Conditional log-likelihood vector}
\label{sec:log-likelihood-vector}

We consider a pre-defined dataset consisting of $N$ prompt-response pairs,
\[
D = \{(x_1,y_1),\dots,(x_N,y_N)\}.
\]
As a vector representation of a language model $p_i$ based on this dataset $D$, we define the log-likelihood vector $\bmell_i$ as
\[
\bmell_i = (\ell_{i1},\dots,\ell_{iN})^\top\in\bbR^N,
\]
where $\ell_{is}$ is a log-likelihood associated with the pair $(x_s,y_s)$.
In our method, $\ell_{is}$ is the log-likelihood assigned to the response $y_s$ by the conditional probability distribution $p_i({\,\cdot\,}| x_s)$ given the input prompt $x_s$, computed using (\ref{eq:piyx}) as
\begin{equation} \label{eq:log-likelihood}
  \ell_{is} = \log p_i(y_s | x_s).
\end{equation}
We call $\bmell_i$ with these components the conditional log-likelihood vector.
Furthermore, by subtracting the mean log-likelihood
\[
\bar\ell_i = \frac{1}{N}\sum_{s=1}^N \ell_{is}
\]
from each component, we define the centered log-likelihood vector
\[
\bmxi_i = (\xi_{i1},\dots,\xi_{iN})^\top\in\bbR^N,
\quad
\xi_{is} = \ell_{is} - \bar\ell_i.
\]
In contrast, the log-likelihood vectors introduced in \citet{modelmap2025} take the same form, but with the components defined by
\begin{equation} \label{eq:unconditional-log-likelihood}
  \ell_{is} = \log p_i(y_s),
\end{equation}
the log-likelihood assigned to $y_s$ by the unconditional probability distribution $p_i(y)$. We refer to these as unconditional log-likelihood vectors in this paper, although \citet{modelmap2025} simply called them log-likelihood vectors. In general, the unconditional definition (\ref{eq:unconditional-log-likelihood}) is appropriate when $y_s$ is a text chunk without preceding context, whereas the conditional definition in (\ref{eq:log-likelihood}) is more appropriate when we are interested in the response $y_s$ to a prompt $x_s$. Figure~\ref{fig:oursetting} illustrates the two settings.

\subsection{Kullback--Leibler divergence}
\label{sec:kl-divergence}

First, for a given prompt $x$, we measure the difference between the conditional probability distributions of two language models $p_i$ and $p_j$ by the conditional KL divergence
\begin{align}
\KL(p_i,p_j\mid x)
= \bbE_{y\sim p_i(y|x)} \left[ \log \frac{p_i(y|x)}{p_j(y|x)} \right].
\label{eq:kl-conditional-given-x}
\end{align}
We assume that the unknown data-generating distribution underlying the dataset $D$ is given by $(x,y)\sim p_0(x,y)=p_0(y|x)p_0(x)$. We refer to $p_0$ as the reference distribution and, in particular, to $p_0(x)$ as the reference prompt distribution.
Then, we consider the average of this conditional KL divergence over the prompt distribution $x\sim p_0(x)$:
\begin{align}
\KL(p_i,p_j) &= \bbE_{x\sim p_0(x)} \Bigl[\KL(p_i,p_j\mid x)\Bigr] \nonumber\\
=& \bbE_{(x,y)\sim p_i(y|x)p_0(x)} \left[ \log \frac{p_i(y|x)}{p_j(y|x)} \right].
\label{eq:kl-conditional}
\end{align}
Here $p_i(y|x)p_0(x)$ is the prompt-response distribution of model $p_i$, that is, its joint distribution over $(x,y)$; the prompt distribution $p_0(x)$ is common to all models.
Furthermore, assuming that both $p_i(y|x)$ and $p_j(y|x)$ are sufficiently close to $p_0(y|x)$, we can apply the argument of \citet{modelmap2025} to the joint distribution over $(x,y)$, so that this KL divergence is approximated by
\begin{align}
\KL(p_i,p_j)\approx \frac{1}{2}\Var_{(x,y)\sim p_0(x,y)} \left[ \log \frac{p_i(y|x)}{p_j(y|x)} \right].
\label{eq:kl-conditional-variance}
\end{align}
Details of the derivation are given in Appendix~\ref{app:kl-variance}.
Approximating this variance by the empirical variance on the dataset $D$, we obtain
\begin{align}
  \KL(p_i,p_j)\approx \frac{1}{2N}\|\bmxi_i - \bmxi_j\|^2
   \label{eq:kl-conditional-xi}
\end{align}
using the centered log-likelihood vectors $\bmxi_i$.
Therefore, the squared Euclidean distance between centered log-likelihood vectors can be interpreted as a quantity that approximates the difference between the conditional probability distributions of language models.
Appendix~\ref{app:conditional-kl-quadratic-approximations} compares the population quantities obtained without centering, with global centering, and with per-prompt centering, while Appendix~\ref{app:empirical-coordinates-sampling-centering} gives their empirical coordinate representations and discusses sampling and the practical role of centering.

\subsection{PMI vector}
\label{sec:pmi-vector}

Recall the conditional log-likelihood vector $\bmell_i$ defined in Section~\ref{sec:log-likelihood-vector}.
Our definition is based on the conditional probability distribution $p_i(y|x)$, whereas \citet{modelmap2025} used the unconditional probability distribution $p_i(y)$. The difference between the two is whether the association between the prompt $x$ and the response $y$ is taken into account. We consider the difference between these two log-likelihood vectors and, because it is closely related to pointwise mutual information, refer to it as the PMI vector for convenience:
\[
\Delta\bmell_i = (\Delta\ell_{i1},\ldots,\Delta\ell_{iN})^\top\in\bbR^N,
\]
where each component is the increase in log-likelihood obtained by taking $x$ into account:
\begin{align}
\Delta\ell_{is} = \log p_i(y_s|x_s) - \log p_i(y_s).
\label{eq:diff-is}
\end{align}
We also define the centered vector
\[
\bmeta_i = (\eta_{i1},\ldots,\eta_{iN})^\top\in\bbR^N
\]
by subtracting the mean $\overline{\Delta\ell}_i = \frac{1}{N}\sum_{s=1}^N \Delta\ell_{is}$ from each component, where $\eta_{is} = \Delta\ell_{is}-\overline{\Delta\ell}_i$. Table~\ref{tab:methods} compares the three representations.

To interpret $\Delta\bmell_i$, we rewrite the right-hand side of (\ref{eq:diff-is}) as
\[
\Delta\ell_{is}
=
\log \frac{p_i(y_s|x_s)p_0(x_s)}{p_i(y_s)p_0(x_s)}.
\]
For the denominator to be the product of the marginals of the prompt-response distribution $p_i(y|x)p_0(x)$ in the numerator, its response term must be
\begin{align}
p_i^{\mathrm{ref}}(y_s)
&:=
\bbE_{x\sim p_0(x)}[p_i(y_s|x)].
\label{eq:reference-marginal}
\end{align}
We call (\ref{eq:reference-marginal}) the reference marginal because it averages over the reference prompt distribution $p_0(x)$. Thus, if $p_i(y_s)$ in (\ref{eq:diff-is}) were given by (\ref{eq:reference-marginal}), $\Delta\ell_{is}$ would be the PMI between $x$ and $y$.

On the other hand, if we consider the joint distribution $p_i(y|x)p_i(x)$ based on the model's own prompt distribution, its response marginal is
\begin{align}
p_i^{\mathrm{model}}(y_s)
&:=
\bbE_{x\sim p_i(x)}[p_i(y_s|x)].
\label{eq:model-marginal}
\end{align}
We call (\ref{eq:model-marginal}) the model marginal. In our experiments, the probability obtained from the language model using the empty-prompt evaluation serves as a practical proxy for this model marginal and is used as $p_i(y_s)$. Appendix~\ref{app:marginal-proxy} examines this substitution by comparing the resulting MI estimates with those obtained from shuffled-prompt Monte Carlo estimation of the reference marginal. The PMI vector is a representation obtained by subtracting the contribution of the unconditional distribution from the conditional log-likelihood vector, and it can be regarded as a feature that more strongly emphasizes information about the dependency structure between prompts and responses.

\begin{figure*}[!t]
\centering
  \includegraphics[width=\linewidth]{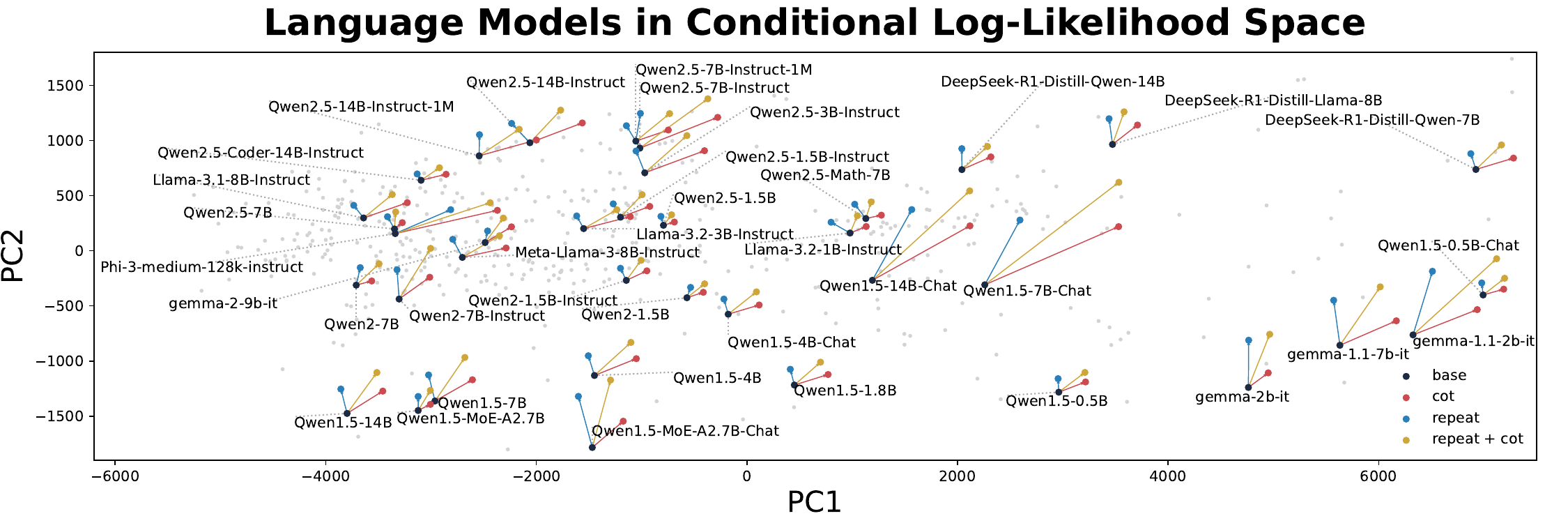}
  \caption{
Visualization of ``prompt shift'' on the model map constructed from the Tulu3 text set.
For each model, we compute conditional log-likelihood vectors for the original prompt \(x\) (base), the prompt with an added CoT phrase \(T_{\mathrm{cot}}(x)\) (cot), the repeated prompt \(T_{\mathrm{rep}}(x)\) (repeat), and the prompt with both transformations applied \(T_{\mathrm{rep}+\mathrm{cot}}(x)\) (repeat+cot).
These vectors are projected onto the linear subspace spanned by the mean shift vectors for \(x \to T_{\mathrm{cot}}(x)\) and \(x \to T_{\mathrm{rep}}(x)\), and visualized by PCA within that subspace. Gray points indicate the base positions, with model names omitted. Systematic shifts induced by prompt transformations and their approximate additive compositionality are clearly observed.
}
  \label{fig:intro}
\end{figure*}

\subsection{Mutual information}
\label{sec:mutual-information}

The extent to which a language model $p_i$ captures the association between a prompt $x$ and a response $y$ can be quantified by mutual information (MI), which is the average of pointwise mutual information. In the theoretical definition below, $p_i(y)$ denotes the reference marginal in (\ref{eq:reference-marginal}):
\begin{align}
\MI(p_i) = \bbE_{(x,y)\sim p_i(y|x)p_0(x)} \left[ \log \frac{p_i(y|x)}{p_i(y)} \right].
\label{eq:mi}
\end{align}

Replacing the reference marginal $p_i(y)$ in (\ref{eq:mi}) with the model marginal in (\ref{eq:model-marginal}), we approximate MI as
\begin{align}
   \MI(p_i) \approx \overline{\Delta\ell}_i.
   \label{eq:mi-bar-delta-ell}
\end{align}
This approximation also replaces the expectation with respect to $(x,y)\sim p_i(y|x)p_0(x)$ in (\ref{eq:mi}) by the empirical average based on $(x,y)\sim p_0(x,y)$.

As a side remark, using the same replacement and by analogy with (\ref{eq:kl-conditional-xi}), MI may be heuristically approximated by
\begin{align}
\MI(p_i)\approx \frac{1}{2N}\|\bmeta_i\|^2,
\label{eq:mi-eta}
\end{align}
which in turn gives an interpretation of the centered PMI vector $\bmeta_i$: the longer this vector, the more mutual information the model captures\footnote{When $p_i(y)$ is the reference marginal in (\ref{eq:reference-marginal}), MI is the KL divergence between the joint distribution $p_i(y|x)p_0(x)$ and the product of marginals $p_i(y)p_0(x)$. Therefore, under the same local assumptions, the variance approximation used for (\ref{eq:kl-conditional-variance}) applies here as well. However, Appendix~\ref{app:marginal-proxy} finds only a weak positive correlation between the MI estimates in Eqs.~(\ref{eq:mi-bar-delta-ell}) and (\ref{eq:mi-eta}), indicating that Eq.~(\ref{eq:mi-eta}) should be regarded only as a rough proxy for MI.}.

\section{Model Maps for Conditional Distributions}
\label{sec:conditional-model-map}

Model maps~\cite{modelmap2025} provide a representation of relationships among language models based on log-likelihood vectors.
This representation can be used not only for visualization but also for approximating distances between models and predicting task performance.
In this section, we introduce two extensions of model maps to conditional distributions: the \textit{conditional model map} defined by conditional log-likelihood vectors and the \textit{PMI model map} defined by PMI vectors. For each experiment, the detailed settings are described in Appendix~\ref{app:details-conditional-model-map}.

\begin{figure*}[!t]
    \centering
    \includegraphics[width=\linewidth]{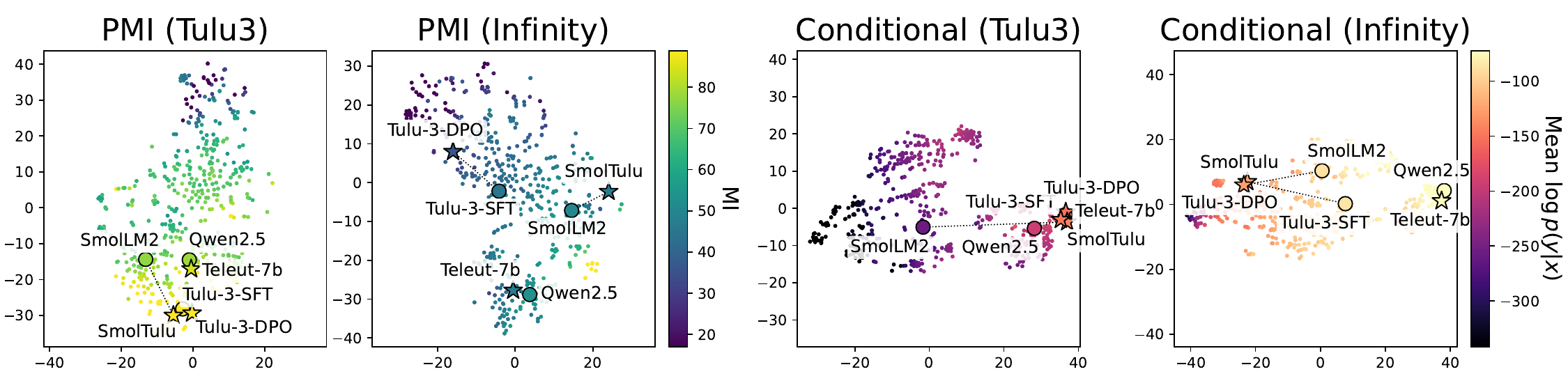}
\caption{\textbf{Comparison of the PMI model map and the conditional model map.} Maps are constructed from PMI vectors (left) and conditional log-likelihood vectors {$\bmell_i$} (right) on Tulu3 and Infinity-Instruct. Colors indicate the mean PMI $\overline{\Delta \ell}_i$ in the PMI maps and the mean conditional log-likelihood $\bar{\ell}_i$ in the conditional maps. Models trained on Tulu3 or Infinity-Instruct are marked with stars, and their corresponding base models are marked with circles. The figure qualitatively illustrates that the PMI model map tends to reflect more clearly the correspondence between the training data and the text set used for visualization.}
    \label{fig:cll_pmi_1}
\end{figure*}

\begin{table*}[t]
\centering
\small
\begin{adjustbox}{width=\linewidth}
\begin{tabular}{llrrrrrrrr}
\toprule
& & \multicolumn{2}{c}{MI} & \multicolumn{2}{c}{$\log p(y|x)$} & \multicolumn{2}{c}{$\Delta$ MI} & \multicolumn{2}{c}{$\Delta \log p(y|x)$} \\
\cmidrule(lr){3-4} \cmidrule(lr){5-6} \cmidrule(lr){7-8} \cmidrule(lr){9-10}
Stage & Model & Tulu3 & Infinity & Tulu3 & Infinity & Tulu3 & Infinity & Tulu3 & Infinity \\
\midrule
Base & \texttt{allenai/Llama-3.1-Tulu-3-8B-SFT} & $97.3$ & $45.3$ & $-125.7$ & $-85.0$ &  &  &  &  \\
Trained (Tulu3) & \texttt{allenai/Llama-3.1-Tulu-3-8B-DPO} & $99.6$ & $38.1$ & $-149.7$ & $-117.0$ & $+2.4$ & $-7.2$ & $-24.0$ & $-32.1$ \\
\midrule
Base & \texttt{HuggingFaceTB/SmolLM2-1.7B-Instruct} & $78.2$ & $50.3$ & $-246.4$ & $-86.3$ &  &  &  &  \\
Trained (Tulu3) & \texttt{SultanR/SmolTulu-1.7b-Instruct} & $112.4$ & $46.5$ & $-138.0$ & $-120.8$ & $+34.3$ & $-3.8$ & $+108.4$ & $-34.5$ \\
\midrule
Base & \texttt{Qwen/Qwen2.5-7B} & $78.5$ & $52.9$ & $-188.5$ & $-54.6$ &  &  &  &  \\
Trained (Tulu3) & \texttt{allura-org/Teleut-7b} & $84.3$ & $46.7$ & $-145.2$ & $-64.5$ & $+5.7$ & $-6.2$ & $+43.3$ & $-9.9$ \\
\midrule
Base & \texttt{Qwen/Qwen2-7B} & $70.1$ & $46.9$ & $-195.6$ & $-60.7$ &  &  &  &  \\
Trained (Infinity) & \texttt{BAAI/Infinity-Instruct-3M-0625-Qwen2-7B} & $68.5$ & $47.3$ & $-190.5$ & $-55.3$ & $-1.6$ & $+0.4$ & $+5.1$ & $+5.4$ \\
\bottomrule
\end{tabular}
\end{adjustbox}
\caption{Mean PMI $\overline{\Delta \ell}_i$ and mean conditional log-likelihood $\bar{\ell}_i$ of selected models on Tulu3 and Infinity-Instruct. For each model pair, the post-trained row additionally reports the improvement over the corresponding base model.}
\label{tab:mi_cond_scores}
\end{table*}

\subsection{Model attributes}
\label{sec:conditional-model-attributes}

Figure~\ref{fig:intro2} shows a conditional model map obtained by visualizing conditional log-likelihood vectors\footnote{The KL approximation in Section~\ref{sec:kl-divergence} is expressed using the centered vectors $\bm{\xi}_i$ in (\ref{eq:kl-conditional-xi}), while Appendices~\ref{app:conditional-kl-quadratic-approximations} and~\ref{app:empirical-coordinates-sampling-centering} show, respectively, that the uncentered representation gives the same second-order approximation at the population level and the same leading accuracy as a finite-sample estimator. We visualize the uncentered vectors $\bm{\ell}_i$ in all figures, following \citet{modelmap2025}. As shown in Eq.~(4) of that paper, the squared distance is decomposed as $\|\bm{\ell}_i-\bm{\ell}_j\|^2=\|\bm{\xi}_i-\bm{\xi}_j\|^2+N(\bar{\ell}_i-\bar{\ell}_j)^2$. Hence, the $\bm{\ell}$-space reflects both the variation captured by $\bm{\xi}_i$ and the variation in $\bar{\ell}_i$.} {$\bmell_i$} computed on the Tulu3 text set using t-SNE.
The top panel shows that models of the same type, such as the Qwen, Mistral, and Llama families, form coherent regions on the map.
In the bottom panel, models located close to each other exhibit similar behavior, and their spatial arrangement shows a certain correspondence with model performance.

\subsection{Prompt shift}
\label{sec:prompt-shift}

Unlike the unconditional setting of \citet{modelmap2025}, conditional log-likelihood vectors allow us to directly examine how the generation distribution of each language model changes on the model map in response to changes in the input prompt.
Let $T(x)$ denote an operation that transforms an input $x$.
In our framework, $p(\cdot|x)$ and $p(\cdot|T(x))$ can be compared in the same space as distinct conditional probability models.
We refer to the systematic movement of the conditional probability distribution in response to a change in the prompt as \textit{prompt shift}.

To investigate prompt shift, we consider three prompt transformations obtained by concatenating text to the original input $x$:
$T_{\mathrm{cot}}(x)=x+c$,
$T_{\mathrm{rep}}(x)=x+x$,
and
$T_{\mathrm{rep+cot}}(x)=x+x+c$.
Here, $c$ is the CoT-inducing phrase ``Let's think step by step.''~\cite{kojima2022large}.
The transformation $T_{\mathrm{rep}}$ corresponds to repeating the prompt twice, which has recently been reported to improve the performance of non-reasoning models~\cite{leviathan2025prompt}.
In the following, we denote the base setting and the settings corresponding to these transformations by
\textsc{base}, \textsc{cot}, \textsc{repeat}, and \textsc{repeat+cot}, respectively.

For each setting, we construct the corresponding dataset $\{(T(x_s),y_s)\}_{s=1}^{N}$ from $\{(x_s,y_s)\}_{s=1}^{N}$ and compute the conditional log-likelihood vector for each model.
The conditional log-likelihood vectors for model $p_i$ under the four settings are denoted by $\bm{\ell}_{\mathrm{base},i}$, $\bm{\ell}_{\mathrm{cot},i}$, $\bm{\ell}_{\mathrm{rep},i}$, and $\bm{\ell}_{\mathrm{rep+cot},i}$, respectively.
We also define three shift vectors $\bm{v}_{\mathrm{cot},i} = \bm{\ell}_{\mathrm{cot},i} - \bm{\ell}_{\mathrm{base},i}$,
$\bm{v}_{\mathrm{rep},i} = \bm{\ell}_{\mathrm{rep},i} - \bm{\ell}_{\mathrm{base},i}$, and
$\bm{v}_{\mathrm{rep+cot},i} = \bm{\ell}_{\mathrm{rep+cot},i} - \bm{\ell}_{\mathrm{base},i}$.

\subsection{Additive compositionality of prompt shift}
\label{sec:addcomp-prompt-shift}

Figure~\ref{fig:intro} visualizes the conditional log-likelihood vectors {$\bmell_i$} under these four settings.
Specifically, for each model, we project the vectors onto the linear subspace spanned by the mean shift vectors $\bar{\bm{v}}_{\mathrm{cot}}$ and $\bar{\bm{v}}_{\mathrm{rep}}$, and then apply PCA within that subspace to obtain a two-dimensional visualization.
Note that the PCA in this projection centers the projected coordinates in the usual sense of PCA; the log-likelihood vectors themselves are not centered.
The black and gray points both represent \textsc{base} models, with model names shown only for the black ones; the gray ones are placed in the background.
The red, blue, and yellow points represent \textsc{cot}, \textsc{repeat}, and \textsc{repeat+cot}, respectively, and are connected by line segments to the corresponding \textsc{base} models.
This figure shows systematic position changes in response to the prompt transformations, indicating prompt shift.

From the positional relationships among these points, an additive compositionality of the shifts induced by \textsc{cot} and \textsc{repeat} is suggested, analogous to the additive compositionality observed in word embeddings~\cite{mikolov2013distributed}.
That is,
\begin{align} \label{eq:v-additive-compositionality}
    \bm{v}_{\mathrm{rep+cot},i} \approx
    \bm{v}_{\mathrm{cot},i} + \bm{v}_{\mathrm{rep},i}
\end{align}
holds.
Indeed, the median and mean of the relative prediction error
\[
\|\bm{v}_{\mathrm{cot},i} + \bm{v}_{\mathrm{rep},i} - \bm{v}_{\mathrm{rep+cot},i}\|/\|\bm{v}_{\mathrm{rep+cot},i} + \bm{c}_i \|
\]
are 0.062 and 0.081, respectively, indicating small errors\footnote{As in \citet{modelmap2025}, we introduce $\bm{c}_i = \bm{\ell}_{\mathrm{base},i} - \bar{\bm{\ell}}_{\mathrm{base}}$ in the denominator in order to evaluate the prediction. For the centered version, we use $\bm{c}_i = \bm{\xi}_{\mathrm{base},i} - \bar{\bm{\xi}}_{\mathrm{base}}$ in the doubly centered model-coordinate space.}. When $\bm{\ell}$ is replaced by $\bm{\xi}$, the corresponding median and mean are 0.074 and 0.095.

Additive compositionality is also observed in the mean conditional log-likelihood $\bar\ell_i$.
Specifically, let $\delta_{\mathrm{cot}}$, $\delta_{\mathrm{rep}}$, and $\delta_{\mathrm{rep+cot}}$ denote the changes in $\bar\ell_i$ from \textsc{base} to \textsc{cot}, \textsc{repeat}, and \textsc{repeat+cot}, respectively.
Then, the Pearson correlation coefficient between $\delta_{\mathrm{rep+cot}}$ and $\delta_{\mathrm{cot}} + \delta_{\mathrm{rep}}$ is 0.987 (bootstrap 95\% confidence interval $[0.959, 0.995]$).

These results suggest that the effects of prompt transformations are composed approximately additively in the conditional log-likelihood vector space\footnote{If the change induced by $T(x)$ is approximated by linearization in the $\bm{\ell}$ space, this property can be understood as a natural consequence of that approximation.}, which opens the possibility of predicting the effect of a composite prompt operation as the sum of the shifts induced by its individual operations.
Furthermore, by transferring a prompt-shift vector obtained from one model to another, it may be possible to predict the effect of an unobserved prompt transformation, as well as the resulting change in task performance.

\subsection{PMI model map and mutual information}
\label{sec:pmi-map}

Table~\ref{tab:mi_cond_scores} reports the mean PMI and the mean conditional log-likelihood $\bar{\ell}$ for Tulu3 and Infinity-Instruct.\footnote{\texttt{Llama-3.1-Tulu-3-8B-DPO} was trained on \texttt{allenai/llama-3.1-tulu-3-8b-preference-mixture}, but we treat it as part of the Tulu3 family for simplicity.} For the mean PMI, a relatively clear tendency is observed: the values increase when the training data and the evaluation text set match, and decrease when they do not. In contrast, such a tendency is not clear for $\bar{\ell}$.
Using models trained on Tulu3 or Infinity-Instruct, we then compare the conditional model map and the PMI model map. Figure~\ref{fig:cll_pmi_1} suggests that, compared with the conditional model map, the PMI model map tends to reflect more clearly the correspondence between the training data and the text set used for visualization. In particular, when the text set used for visualization matches the training data, models trained on that data tend to be located relatively close to one another, whereas when they do not match, such clustering tends to weaken and the configurations become more dispersed.

\section{Task Performance Prediction}
\label{sec:score-prediction}

In Section~\ref{sec:addcomp-prompt-shift}, we demonstrated the additive compositionality of prompt shifts in the log-likelihood vector space. In this section, we investigate whether this structural property transfers to the prediction of downstream task performance. 

\subsection{Experimental setup}
See Appendix~\ref{sec:experimental-settings} for the text sets, the language models, and the computation of the log-likelihood.

\paragraph{Evaluation.}
We measure task performance with the lm-evaluation-harness on five tasks: ARC-Challenge, ARC-Easy, OpenBookQA, PIQA, and TruthfulQA. All evaluations are conducted in a zero-shot setting (\texttt{num\_fewshot}=0) with \texttt{apply\_chat\_template=True}. 

\paragraph{Regression model.}
For each setting $T \in \{\mathrm{base,cot,rep,rep{+}cot}\}$, we collect the centered log-likelihood vectors into the matrix $\bm{\Xi}_T = [\bm{\xi}_{T,1},\ldots,\bm{\xi}_{T,K}]^{\top} \in \bbR^{K \times N}$. We concatenate $\{\bm{\Xi}_T\}$ and the corresponding score matrices row-wise, and train a ridge regression model $\hat{f}$, which maps $\bm{\xi}_{T,i}$ to the five task scores (RidgeCV, $\alpha \in [10^{-3}, 10^{5}]$, 5-fold CV), using an 8:2 model-wise train-test split ($K_{\mathrm{train}}=388$, $K_{\mathrm{test}}=97$). 
The optimal $\alpha$ is $28{,}072$. The analysis is conducted on the $K=485$ models that successfully completed task evaluations across all four conditions.
Confidence intervals are 95\% percentile bootstrap intervals obtained by resampling the held-out test models with replacement ($B{=}2{,}000$) and recomputing the correlations, keeping the fitted regression model fixed; for the additivity correlation in Section~\ref{sec:addcomp-prompt-shift}, all $K{=}552$ models are resampled instead.

\subsection{Experimental results}
\paragraph{Additive compositionality at the score level.}
First, without using a regression model, we check if the task score changes themselves exhibit additive properties. For each model $p_i$, we compare the sum of individual score changes $\Delta s_{\mathrm{cot},i} + \Delta s_{\mathrm{rep},i}$ with the actual score change induced by \textsc{repeat+cot}, $\Delta s_{\mathrm{rep+cot},i}$. 
Across $K=485$ models, the Pearson correlation coefficients range from 0.837 (TruthfulQA) to 0.923 (ARC-Easy), indicating that the additive compositionality observed in the log-likelihood space (Section~\ref{sec:addcomp-prompt-shift}) directly translates to task score changes.

\paragraph{Task score prediction.}
Next, we examine whether the centered conditional log-likelihood vectors $\bm{\xi}$ can predict task scores. As shown in Table~\ref{tab:score-pred}, they are reliable predictors of downstream task performance, achieving both $r$ and $\rho$ of at least $0.82$ for all tasks except OpenBookQA ($r=0.586$, $\rho=0.797$).
We also observed that the conditional log-likelihood vector performed slightly better than predictions based on the unconditional log-likelihood vector. The gain is small, however, and the main merit of the conditional representation lies in enabling analyses of prompt-induced changes, such as the prompt shifts in Section~\ref{sec:prompt-shift}. See Appendix~\ref{app:unconditional-baseline} for details.

\begin{table}[!t]
\centering
\begin{adjustbox}{width=\linewidth}
\begin{tabular}{lcc}
\toprule
Task & Pearson $r$ & Spearman $\rho$ \\
\midrule
ARC-Challenge   & 0.829 [0.783, 0.867] & 0.821 [0.765, 0.869] \\
ARC-Easy        & 0.841 [0.803, 0.872] & 0.824 [0.778, 0.863] \\
OpenBookQA      & 0.586 [0.363, 0.807] & 0.797 [0.728, 0.854] \\
PIQA            & 0.906 [0.881, 0.931] & 0.945 [0.926, 0.960] \\
TruthfulQA      & 0.842 [0.790, 0.890] & 0.869 [0.820, 0.906] \\
\bottomrule
\end{tabular}
\end{adjustbox}
\caption{Correlation between predicted and true task scores on the test set, using Ridge regression based on conditional log-likelihood vectors. The input is concatenated across the four prompt conditions, with an 8:2 model-wise split. Brackets denote bootstrap 95\% confidence intervals.}
\label{tab:score-pred}
\end{table}

\begin{table}[!t]
\centering
\begin{adjustbox}{width=\linewidth}
\begin{tabular}{lccc}
\toprule
 & \multicolumn{3}{c}{Pearson $r$} \\
\cmidrule(lr){2-4}
Task & $\Delta$\textsc{cot} & $\Delta$\textsc{repeat} & $\Delta$\textsc{repeat+cot} \\
\midrule
ARC-Challenge & 0.694 [0.541, 0.827] & 0.678 [0.567, 0.773] & 0.588 [0.366, 0.778] \\
ARC-Easy      & 0.728 [0.625, 0.817] & 0.359 [0.155, 0.540] & 0.603 [0.448, 0.735] \\
OpenBookQA    & 0.595 [0.459, 0.701] & 0.250 [$-0.021$, 0.500] & 0.501 [0.323, 0.662] \\
PIQA          & 0.678 [0.529, 0.794] & 0.408 [0.135, 0.610] & 0.518 [0.290, 0.702] \\
TruthfulQA    & 0.486 [0.327, 0.648] & 0.557 [0.405, 0.691] & 0.721 [0.624, 0.799] \\
\bottomrule
\end{tabular}
\end{adjustbox}
\caption{Correlation (Pearson $r$) between predicted and true score changes induced by each prompt transformation on the test set. Brackets denote bootstrap 95\% confidence intervals.}
\label{tab:delta-score-pred}
\end{table}

\paragraph{Predicting score changes via shift vectors.}
We further investigate whether the regression model can predict the score changes induced by prompt transformations. We use the difference between the predicted score for condition $T$ and the predicted score for the \textsc{base} condition, $\hat{f}(\bm{\xi}_{T,i}) - \hat{f}(\bm{\xi}_{\mathrm{base},i})$, as the predicted score change. 
As shown in Table~\ref{tab:delta-score-pred}, the prediction performance varies depending on the condition and task ($r \approx 0.25$ to $0.73$), and the bootstrap 95\% confidence intervals exclude zero for all but one condition-task pair. These results suggest that prompt shifts in the log-likelihood space retain information about both the direction and magnitude of performance changes.

\paragraph{Predicting the effect of composed prompts.}
Finally, leveraging the additive compositionality in Eq.~(\ref{eq:v-additive-compositionality}), we examine whether we can predict the task score for the \textsc{repeat+cot} condition without directly observing its corresponding log-likelihoods.
We compute the synthetic centered log-likelihood vector $\hat{\bm{\xi}}_{\mathrm{rep+cot},i} = \bm{\xi}_{\mathrm{cot},i} + \bm{\xi}_{\mathrm{rep},i} - \bm{\xi}_{\mathrm{base},i}$ and feed it into our trained regression model as a proxy for \textsc{repeat+cot}.
Test-set Pearson correlations are high ($0.819$–$0.913$) for all tasks except OpenBookQA ($0.418$), for which the Spearman rank correlation nevertheless remains strong at $0.703$. These results demonstrate that additive compositionality in the log-likelihood space can be leveraged to approximately predict the performance impact of composite prompt operations. We also conducted experiments using log-likelihood vectors obtained from two additional datasets and observed consistent results. See Appendix~\ref{app:additional-text-sets} for details.

\begin{figure}[t]
  \centering
  \begin{subfigure}[t]{0.485\columnwidth}
    \centering
    \includegraphics[width=\linewidth]{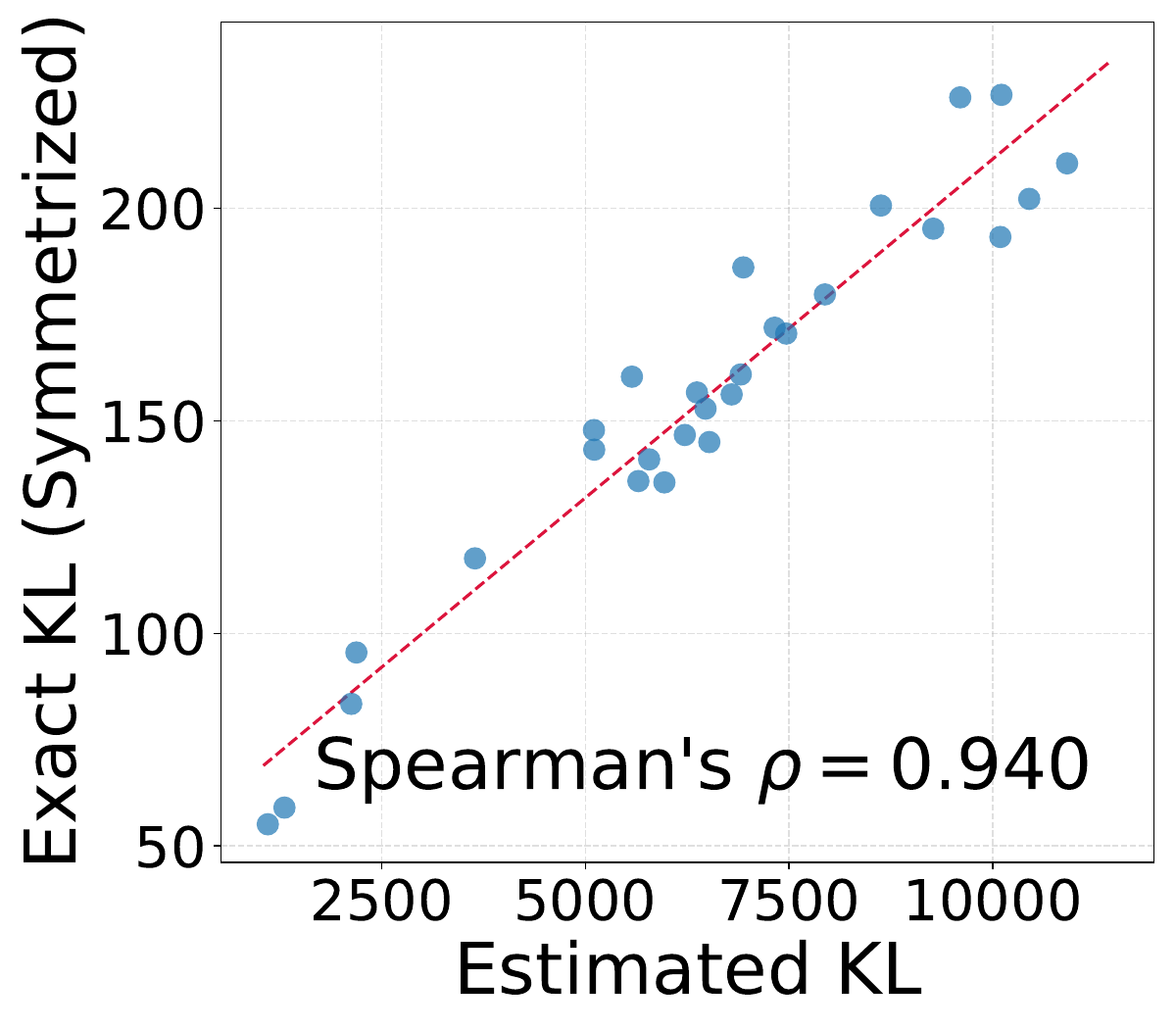}
    \caption{Comparison of KL estimates obtained from the squared Euclidean distance between globally centered log-likelihood vectors and from the Monte Carlo approximation to the definition of KL divergence.}
    \label{fig:kl_vs_pseudo_kl_scatter}
  \end{subfigure}
  \hfill
  \begin{subfigure}[t]{0.485\columnwidth}
    \centering
    \includegraphics[width=\linewidth]{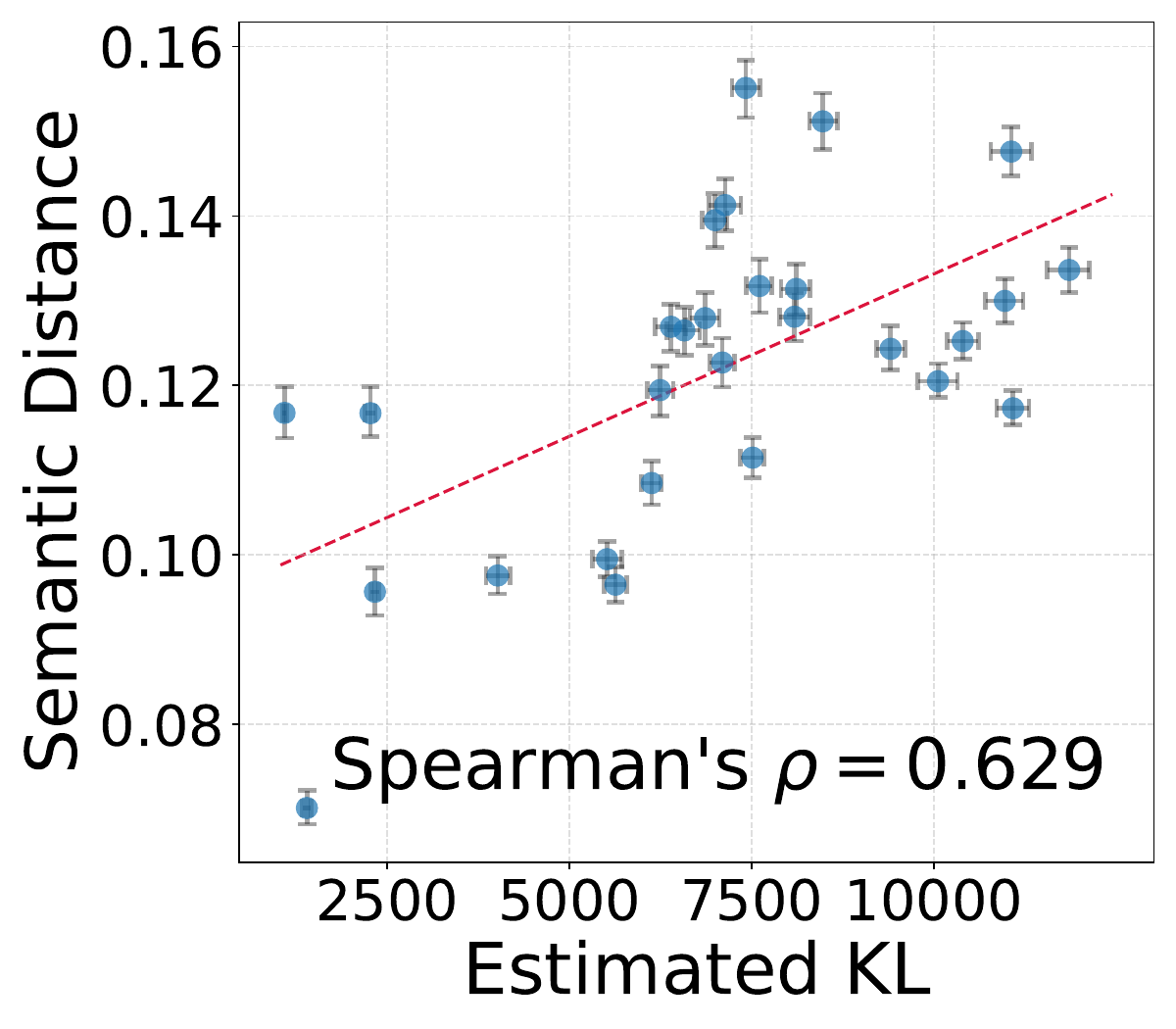}
    \caption{Comparison between the KL divergence estimated from log-likelihood vectors and the semantic distance based on text embeddings. Error bars indicate bootstrap 95\% confidence intervals.}
    \label{fig:semantic_scatter}
  \end{subfigure}
\caption{Empirical comparisons of model distances.}
  \label{fig:section6_combined}
\end{figure}

\section{Empirical Validation}
\label{sec:empirical-validation}

In this section, we empirically examine whether distances based on log-likelihood vectors provide a useful approximation to model differences. Our main comparisons use $K'=8$ instruction-tuned language models and $N=10{,}000$ prompts sampled from the Tulu3 dataset. For the comparison with Monte Carlo estimates of KL divergence, we additionally conduct an experiment with $K'=24$ models and $N=1{,}000$ prompts, as reported in Appendix~\ref{app:24models}. See Appendix~\ref{app:details-empirical-validation} for the list of the eight models and other details of the main experiments.

\subsection{Comparison with Monte Carlo estimates of KL divergence}
\label{sec:monte-carlo-kl}

For each prompt $x_s$, we generate one response $y_s^i \sim p_i(\cdot|x_s)$ from each model $p_i$. Using these generated responses, we estimate
$\mathrm{KL}(p_i,p_j)$ defined in (\ref{eq:kl-conditional}) by the Monte Carlo approximation
\[
\mathrm{KL}(p_i,p_j)
\approx
\frac{1}{N}\sum_{s=1}^{N}
\log \frac{p_i(y_s^i|x_s)}{p_j(y_s^i|x_s)}.
\]
We then compare this Monte Carlo estimate with the KL estimate based on Eq.~(\ref{eq:kl-conditional-xi}), which is symmetric in the two models.
Because the Monte Carlo estimate is directional, we symmetrize it before the comparison shown in Figure~\ref{fig:kl_vs_pseudo_kl_scatter}; see Appendix~\ref{app:details-empirical-validation} for the definition.

\paragraph{Results.}
The left panel of Figure~\ref{fig:section6_combined} shows a strong rank correlation between the two estimates (Spearman's $\rho=0.94$), providing empirical support for the log-likelihood-vector-based approximation. The 24-model experiment in Appendix~\ref{app:24models} yields a lower but still substantial rank correlation (Spearman's $\rho=0.83$).

\subsection{Comparison with semantic distance}
\label{sec:semantic-distance}

We next compare the KL divergence estimated from log-likelihood vectors with an embedding-based semantic distance. For each prompt $x_s$ and each model $p_i$, we generate $M=10$ responses. Following \citet{jiang2025artificial}, we average pairwise cosine similarities between responses from two models and define the semantic distance as the complement of this average; see Appendix~\ref{app:details-empirical-validation} for details.
The right panel of Figure~\ref{fig:section6_combined} plots the semantic distance against the KL divergence estimated from log-likelihood vectors. The two quantities show a positive association. Bootstrap confidence intervals also indicate that the log-likelihood-vector-based estimate is at least comparable in stability to the embedding-based semantic distance.

\section{Conclusion}

We proposed a framework for constructing model maps for prompt-response distributions using conditional log-likelihood vectors. In the resulting vector space, squared Euclidean distances between models approximate the KL divergence between their conditional distributions averaged over the reference prompt distribution, extending previous log-likelihood-vector-based model maps to conditional settings. We also introduced PMI vectors to reduce the contribution of the response marginal; in some cases, the resulting maps more clearly reflect training-data-related differences. We further showed that prompt modifications induce systematic shifts in model positions and that these shifts exhibit approximate additive compositionality. We also used the vectors to predict downstream task performance and leveraged their additive structure to approximate the effects of composite prompt operations without directly observing the corresponding log-likelihoods. The resulting distance estimates showed strong rank correlations with Monte Carlo KL estimates and a positive association with embedding-based semantic distance. Overall, the framework supports the comparison, analysis, and prediction of input-dependent behavior across language models.

\section*{Limitations}

\begin{itemize}

    \item This study focuses on models whose weights are publicly available, and extending the method to closed models remains future work.

    \item The experiments are conducted mainly on instruction-style prompt-response datasets such as Tulu3 and Infinity-Instruct. It remains unclear to what extent the proposed framework generalizes to other settings, such as non-instructional text, multilingual data, code generation, or longer and more complex interactions.

    \item The effectiveness of PMI-based quantities is not uniform across all analyses. Although PMI vectors can in some cases reflect training-data-related differences more clearly than conditional log-likelihood vectors, this tendency is not consistently observed, and the practical advantage of PMI-based model maps may depend on the dataset and models being compared.

    \item The prompt-shift analysis is limited to a small number of prompt transformations, namely appending a CoT suffix and repeating the prompt. It remains an open question whether similar geometric regularities and approximate additive compositionality hold for a broader class of prompt edits.

    \item The visualizations in this paper rely on low-dimensional projections such as t-SNE and PCA. While these are useful for qualitative inspection, they do not fully preserve the geometry of the original high-dimensional space, so apparent proximity or clustering in two dimensions should be interpreted with caution.

    \item The empirical validation is conducted on limited sets of language models: eight models in the main comparisons in Section~\ref{sec:empirical-validation} and 24 models in the additional comparison with Monte Carlo estimates of KL divergence in Appendix~\ref{app:24models}. Although the observed correlations support the proposed approximation, the robustness of these findings across a broader range of models, decoding settings, and evaluation prompts remains to be investigated.

    \item The proposed framework depends on specific formatting choices for prompt-response pairs, including prompt templates and prompt transformations. Such implementation details may affect the resulting log-likelihood vectors and model positions, and disentangling these effects from more intrinsic behavioral differences remains future work.
    
    \item For all KL estimates derived from log-likelihood vectors, we apply global centering, whereas the model maps in Figures~\ref{fig:intro2}, \ref{fig:intro}, and~\ref{fig:cll_pmi_1} visualize uncentered vectors; per-prompt centering is not used in any of our experiments. We therefore do not empirically compare the representations obtained without centering, with global centering, and with per-prompt centering. Empirically assessing how the corresponding retained or removed components affect the resulting model distances remains future work.

    \item The KL-divergence approximation in (\ref{eq:kl-conditional-xi}) and the second-order equivalence among the representations obtained without centering, with global centering, and with per-prompt centering in Appendices~\ref{app:conditional-kl-quadratic-approximations} and~\ref{app:empirical-coordinates-sampling-centering} rely on a local approximation around $p_0$. The variance decompositions themselves remain exact, but when the models are not sufficiently close to $p_0$, the $O(\lambda^3)$ terms need not be negligible and the three representations are not guaranteed to approximate KL divergence accurately.

    \item Each component $\Delta\ell_{is}$ of the PMI vector represents the relationship between the prompt and the response, but interpreting it as PMI requires the reference marginal in (\ref{eq:reference-marginal}). The main experiments instead use an empty-prompt estimate of the model marginal in (\ref{eq:model-marginal}). Appendix~\ref{app:marginal-proxy} compares the resulting MI estimates with those obtained using a reference-marginal estimate based on $C=10$ shuffled-prompt permutations for a subset of 26 models on Tulu3. Although the resulting positive correlations support the empty-prompt estimate as a practical substitute for model-level comparison, this limited experiment does not imply that the two marginals agree or that the same relationship holds for other model sets, text sets, or larger values of $C$.

    \item For (\ref{eq:mi-eta}) to be a good approximation, both $p_i(y|x)p_0(x)$ and $p_i(y)p_0(x)$, where $p_i(y)$ is the reference marginal in (\ref{eq:reference-marginal}), must be sufficiently close to $p_0(y|x)p_0(x)$. When responses depend strongly on prompts, these two distributions are generally not close, so this assumption may fail. Thus, (\ref{eq:mi-eta}) should be regarded only as a rough proxy for MI.

    \item The estimated mutual information $\MI(p_i)$ measures the strength of association between the prompt $x$ and the response $y$. However, a strong association between $x$ and $y$ does not necessarily imply that the response is appropriate.
\end{itemize}

\section*{Acknowledgments}

This study was partially supported by JSPS KAKENHI JP22H05106, JP23K28045, JP26H02495 (to HS), JP26KJ1485 (to MO), JST CREST JPMJCR21N3 (to HS), JST Moonshot JPMJMS263I (to HS), and JST BOOST JPMJBS2407 (to YT, MO).

\appendix
\numberwithin{equation}{section}

\section{Related Work}
\label{sec:related}

Existing studies comparing a large number of language models can be broadly categorized, according to the type of information they use, into approaches based on model internals, generated texts, and log-likelihood vectors.

\paragraph{Comparison based on model internals.}
Comparisons based on model weights and intermediate representations can directly examine similarities in internal structure. For example, \citet{zhu2025independence} analyzed relationships among models based on statistical similarities of their parameters, and \citet{horwitz2025we} compared the weights of many models on Hugging Face to reveal relationships that are not easily visible from publicly available information alone. In addition, \citet{zhou-etal-2025-linguistic} measured similarities among models using changes in activations induced by linguistic minimal pairs. 
While these approaches are useful for analyzing model internals, they do not readily provide a common basis for comparison across models with different architectures or representation spaces.

\paragraph{Comparison based on generated texts.}
Comparisons based on generated texts can directly address the practical setting of conditional generation. \citet{jiang2025artificial} mapped generated responses into a text embedding space and measured differences among models based on distances in that space. In addition, \citet{yax2025phylolm} introduced a similarity measure based on conditional probabilities and estimated phylogenetic distances among language models. 
However, such approaches can be affected by the randomness of the generation process and by the choice of embedding-based representations for generated outputs.

\paragraph{Comparison based on log-likelihood vectors.}
Methods that represent models as vectors based on their log-likelihoods on a pre-defined set of texts~\cite{modelmap2025, oyama-etal-2025-likelihood, kishino-etal-2026-establishing} are useful in that they enable many models to be compared in a shared space without relying on generation, and they also allow distances between vectors to be interpreted in relation to the KL divergence between probability distributions. However, existing methods mainly focus on unconditional distributions and do not directly address the setting of conditional generation.

Our study aims to fill this gap. Specifically, we introduce comparisons based on conditional log-likelihood vectors and use PMI-vector representations that subtract unconditional components, thereby providing a framework for comparing the conditional probability distributions of many language models.

\section{Experimental Settings}
\label{sec:experimental-settings}

In this section, we describe the text sets used in our experiments, the language models selected for analysis, and the procedure for computing the log-likelihood vectors. 

\subsection{Selection of text data}
\label{sec:setting-text-dataset}

In the main experiments, we use two text sets, each denoted by $D=\{(x_s,y_s)\}_{s=1}^{N}$.
The first consists of $N=10{,}000$ prompt-response pairs extracted from \texttt{tulu-3-sft-mixture}~\cite{lambert2025tulu}\footnote{\url{https://huggingface.co/datasets/allenai/tulu-3-sft-mixture}}.
The second consists of $N=10{,}000$ prompt-response pairs extracted from \texttt{Infinity-Instruct-7M}~\cite{li2025infinity}\footnote{\url{https://huggingface.co/datasets/BAAI/Infinity-Instruct}}.
For each of these two text sets, we construct a model map and conduct experiments on mutual information in order to analyze how the observed tendencies differ across text sets.
Appendix~\ref{app:additional-text-sets} additionally evaluates robustness to the choice of text set using $N=5{,}000$ prompt-response pairs from each of Tulu3, Infinity-Instruct, and UltraChat~\citep{ding-etal-2023-enhancing}.

\subsection{Selection of language models}
\label{sec:setting-language-models}

To construct the model maps, we consider text generation models listed on Open LLM Leaderboard v2.
Among them, we restrict our analysis to models with at least 1,000 downloads and an available \texttt{chat\_template}.
We also restrict the model size to no larger than 70B due to computational resource constraints.
In addition, for each model, we attempt to compute log-likelihoods and retain only those for which the computation is numerically stable, without overflow or related issues.
The resulting number of models differs across experiments; specifically, we use $K=552$ models for the conditional model map and prompt-shift analyses, $K=481$ models for the PMI model map analysis, and $K=485$ models for the task score prediction in Section~\ref{sec:score-prediction}. The empirical validation in Section~\ref{sec:empirical-validation} uses $K'=8$ instruction-tuned models, and the additional experiments in Appendices~\ref{app:marginal-proxy}, \ref{app:additional-text-sets}, \ref{app:unconditional-baseline}, and~\ref{app:24models} use $K'=26$, $K=100$, $K=438$, and $K'=24$ models, respectively.

\subsection{Computation of the log-likelihood}
\label{sec:setting-log-likelihood-computation}

For each language model, we compute the conditional log-likelihood $\log p(y|x)$ and the unconditional log-likelihood $\log p(y)$ for each text pair $(x,y)$.
To compute $\log p(y|x)$, we feed each prompt-response pair into the model using the corresponding \texttt{chat\_template}.
We obtain $\log p(y)$, the empty-prompt estimate of the model marginal in Eq.~(\ref{eq:model-marginal}), using the same \texttt{chat\_template} with the user input left empty.
All computations are performed using bfloat16 precision.
In addition, to reduce the influence of extremely small log-likelihood values, we apply clipping separately to $\log p(y|x)$ and $\log p(y)$ using the lower 2\% threshold computed from all model-text combinations.

\section{Details of the Variance Representation of KL Divergence}
\label{app:kl-variance}

In this section, we derive the variance representation of KL divergence in (\ref{eq:kl-conditional-variance}) and the corresponding approximation in (\ref{eq:kl-conditional-xi}).
First, we transform the right-hand side of (\ref{eq:kl-conditional}) as follows:
\begin{align*}
&\KL(p_i,p_j) \\
=& \bbE_{(x,y)\sim p_i(y|x)p_0(x)} \left[ \log \frac{p_i(y|x)}{p_j(y|x)} \right] \\
=& \bbE_{(x,y)\sim p_i(y|x)p_0(x)} \left[ \log \frac{p_i(y|x)p_0(x)}{p_j(y|x)p_0(x)} \right].
\end{align*}
This expression represents the KL divergence between the joint distributions $p_i(y|x)p_0(x)$ and $p_j(y|x)p_0(x)$ over $(x,y)$.
We assume that both $p_i(y|x)$ and $p_j(y|x)$ are sufficiently close to $p_0(y|x)$. Then, the corresponding joint distributions $p_i(y|x)p_0(x)$ and $p_j(y|x)p_0(x)$ can also be regarded as sufficiently close to $p_0(y|x)p_0(x)=p_0(x,y)$. Therefore, we can apply Eq.~(2) of \citet{modelmap2025} to the joint distribution over $(x,y)$, obtaining
\begin{align}
2\KL(p_i,p_j)
\approx{}& \Var_{(x,y)\sim p_0(x,y)} \left[ \log \frac{p_i(y|x)p_0(x)}{p_j(y|x)p_0(x)} \right] \nonumber\\
&\hspace{-2em}= \Var_{(x,y)\sim p_0(x,y)} \left[ \log \frac{p_i(y|x)}{p_j(y|x)} \right].
\label{eq:kl-joint-variance}
\end{align}
This yields (\ref{eq:kl-conditional-variance}).
To approximate this variance from the dataset $D$, we use the empirical variance:
\begin{align}
2\KL(p_i,p_j)& \approx \frac{1}{N}\sum_{s=1}^N \biggl[ \log \frac{p_i(y_s|x_s)}{p_j(y_s|x_s)} - \nonumber\\
&\qquad \qquad \frac{1}{N}\sum_{s'=1}^N \log \frac{p_i(y_{s'}|x_{s'})}{p_j(y_{s'}|x_{s'})} \biggr]^2 \nonumber\\
=& \frac{1}{N}\sum_{s=1}^N \Bigl( \xi_{is} - \xi_{js} \Bigr)^2 \nonumber\\
=& \frac{1}{N}\|\bmxi_i - \bmxi_j\|^2.
\label{eq:kl-global-coordinate-m1}
\end{align}
This yields (\ref{eq:kl-conditional-xi}).

\section{Quadratic Approximations of Conditional KL Divergence}
\label{app:conditional-kl-quadratic-approximations}

Equation~(\ref{eq:kl-conditional-variance}), derived in Appendix~\ref{app:kl-variance}, uses a globally centered variance, whereas the model maps visualize uncentered log-likelihood vectors.
Appendix~\ref{app:empirical-coordinates-sampling-centering} considers the more general response-sampling setup in (\ref{eq:multiple-responses-per-prompt}), in which $M$ conditionally independent responses are drawn for each prompt.
In this setting, the conditional variance at each prompt can be estimated, motivating per-prompt centering.
We derive exact population relations among these three centering schemes and show their second-order equivalence; Appendix~\ref{app:empirical-coordinates-sampling-centering} gives their sample coordinates.
For notational convenience, we work on the scale of twice the KL divergence throughout Appendices~\ref{app:conditional-kl-quadratic-approximations} and~\ref{app:empirical-coordinates-sampling-centering}.

\subsection{Three quadratic quantities}

Unless otherwise stated, expectations and variances in this appendix are taken with respect to
\[
(x,y)\sim p_0(x,y)=p_0(y|x)p_0(x).
\]
We consider the log-likelihood ratio
\[
r_{ij}(x,y)
:=
\log p_i(y|x)-\log p_j(y|x).
\]
Define the conditional and overall expectations of the log-likelihood ratio by
\[
\begin{aligned}
\mu_{ij}(x)
&:=
\bbE_{y\sim p_0(y|x)}[r_{ij}(x,y)],\\
\mu_{ij}
&:=
\bbE_{(x,y)\sim p_0(x,y)}[r_{ij}(x,y)],
\end{aligned}
\]
and let
\[
V_{ij}
:=
\Var_{x\sim p_0(x)}\!\left(\mu_{ij}(x)\right).
\]
The three centering schemes give the following population quadratic quantities:
\begin{align}
\mathcal{A}_{\text{uncentered}}
&:=
\bbE_{(x,y)\sim p_0(x,y)}[r_{ij}(x,y)^2],\nonumber\\
\mathcal{A}_{\text{global}}
&:=
\Var_{(x,y)\sim p_0(x,y)}[r_{ij}(x,y)],\nonumber\\
\mathcal{A}_{\text{per-prompt}}
&:=
\bbE_{x\sim p_0(x)}\!
\left[
\Var_{y\sim p_0(y|x)}[r_{ij}(x,y)]
\right].
\label{eq:kl-three-quadratic-quantities}
\end{align}
The global quantity $\mathcal{A}_{\text{global}}$ is the variance on the last line of (\ref{eq:kl-joint-variance}), so Appendix~\ref{app:kl-variance} gives $2\KL(p_i,p_j)\approx\mathcal{A}_{\text{global}}$.
The uncentered quantity $\mathcal{A}_{\text{uncentered}}$ is the population counterpart of the squared distance between uncentered log-likelihood vectors used in model maps \citep{Shimodaira1993modelmap,shimodaira1998graphical}.
The per-prompt quantity $\mathcal{A}_{\text{per-prompt}}$ arises when the local KL argument of \citet{modelmap2025} is applied to the conditional response distribution $y\sim p_0(y|x)$ at each prompt $x$ and then averaged over $x\sim p_0(x)$.

\subsection{Exact relations among the three quantities}

For brevity, hereafter we omit explicit distributions from expectation and variance subscripts whenever they are implied by $(x,y)\sim p_0(x,y)$; for example, $\bbE_x[\cdot]$, $\Var_x(\cdot)$, and $\bbE[\cdot|x]$ denote $\bbE_{x\sim p_0(x)}[\cdot]$, $\Var_{x\sim p_0(x)}(\cdot)$, and $\bbE_{y\sim p_0(y|x)}[\cdot]$, respectively.
With this notation, the second-moment decomposition and the law of total variance give
\begin{align}
\mathcal{A}_{\text{uncentered}}
&=
\mathcal{A}_{\text{global}}
+\mu_{ij}^2,\nonumber\\
\mathcal{A}_{\text{global}}
&=
\mathcal{A}_{\text{per-prompt}}
+V_{ij},\nonumber\\
\mathcal{A}_{\text{uncentered}}
&=
\mathcal{A}_{\text{per-prompt}}
+\bbE_x[\mu_{ij}(x)^2].
\label{eq:kl-centering-decomposition}
\end{align}
These identities are exact and do not require the models to be close to $p_0$.
In particular,
\[
\mathcal{A}_{\text{uncentered}}
\geq
\mathcal{A}_{\text{global}}
\geq
\mathcal{A}_{\text{per-prompt}}.
\]
The first gap in (\ref{eq:kl-centering-decomposition}), $\mu_{ij}^2$, is the squared overall mean of the pairwise log-likelihood ratio.
The second gap, $V_{ij}$, is the variation across prompts in its conditional mean.
Thus, global centering removes the overall mean component, whereas per-prompt centering additionally removes the between-prompt variation in the conditional mean.

\subsection{Local equivalence to conditional KL divergence}

We now relate the exact decomposition above to conditional KL divergence using the local setting of \citet{modelmap2025} around $p_0$.
For $k=i,j$, let
\[
d_k(x,y)
:=
\log p_k(y|x)-\log p_0(y|x),
\]
so that $r_{ij}=d_i-d_j$.
We assume common support and
\[
\begin{gathered}
d_k(X,Y)=O_p(\lambda),\\
(X,Y)\sim p_0,
\quad
\lambda\to0.
\end{gathered}
\]
We also assume sufficient conditional moment bounds so that the Taylor remainders are $O_p(\lambda^3)$ and the stochastic orders below may be squared and averaged over $X$.

Because $p_k(\cdot|X)$ is normalized for every $X$,
\[
\bbE
\left[
e^{d_k(X,Y)}
\mid X
\right]
=1.
\]
Expanding the exponential therefore gives
\begin{align}
\bbE[d_k(X,Y)|X]
={}&
-\frac12
\bbE[d_k(X,Y)^2|X]\nonumber\\
&+O_p(\lambda^3)
=
O_p(\lambda^2).
\label{eq:kl-local-conditional-mean}
\end{align}
Consequently,\footnote{The $O_p(\lambda^2)$ conclusion in (\ref{eq:kl-local-conditional-mean}) can alternatively be obtained from the identity $\bbE[d_k(X,Y)|X]=-\KL(p_0,p_k | X)$. Since conditional KL divergence is locally quadratic in the same setting, this identity gives $\bbE[d_k(X,Y)|X]=O_p(\lambda^2)$.}
\[
\mu_{ij}(X)
=
\bbE[d_i-d_j|X]
=
O_p(\lambda^2).
\]
Under the stated moment conditions,
\[
V_{ij}=O(\lambda^4),
\quad
\mu_{ij}^2=O(\lambda^4),
\]
while each quantity in (\ref{eq:kl-three-quadratic-quantities}) is $O(\lambda^2)$.
Thus, the three quantities are approximately equal to leading second order,
with differences only of order $O(\lambda^4)$:
\[
\mathcal{A}_{\text{uncentered}}
\approx
\mathcal{A}_{\text{global}}
\approx
\mathcal{A}_{\text{per-prompt}}.
\]
Applying the local expansion of \citet{modelmap2025}, including its $O(\lambda^3)$ remainder, to the joint distributions used in Appendix~\ref{app:kl-variance} gives
\[
2\KL(p_i,p_j)
=
\mathcal{A}_{\text{global}}
+O(\lambda^3).
\]
Combining this result with the exact gaps in (\ref{eq:kl-centering-decomposition}) yields
\begin{align}
2\KL(p_i,p_j)
&=
\mathcal{A}_{\text{uncentered}}+O(\lambda^3),\nonumber\\
2\KL(p_i,p_j)
&=
\mathcal{A}_{\text{global}}+O(\lambda^3),\nonumber\\
2\KL(p_i,p_j)
&=
\mathcal{A}_{\text{per-prompt}}+O(\lambda^3).
\label{eq:kl-three-representations}
\end{align}
Thus, although the three quantities are not exactly equal, their differences are only $O(\lambda^4)$.
This fourth-order discrepancy lies beyond the $O(\lambda^3)$ error of the local KL approximation.
Hence, they share the same leading $O(\lambda^2)$ term.

\section{Empirical Coordinates, Sampling, and Centering}
\label{app:empirical-coordinates-sampling-centering}

Appendix~\ref{app:conditional-kl-quadratic-approximations} showed that the three population quantities differ only by $O(\lambda^4)$.
We now construct their empirical counterparts, connect them to (\ref{eq:kl-conditional-xi}) and the uncentered vectors used in our figures, and discuss sampling and centering.

\subsection{Empirical coordinate representations}

For positive integers $N$ and $M$, draw
\[
x_1,\ldots,x_N
\overset{\mathrm{i.i.d.}}{\sim}
p_0(x),
\]
and, conditional on these prompts, independently draw $M$ responses for each $s$:
\begin{equation}
y_{s1},\ldots,y_{sM}
\overset{\mathrm{i.i.d.}}{\sim}
p_0(y|x_s).
\label{eq:multiple-responses-per-prompt}
\end{equation}
Responses drawn for the same prompt are conditionally independent but are generally dependent marginally because they share $x_s$.
The dataset $D$ used in this paper has $M=1$, but we retain general $M$ to make the three empirical representations explicit.

Fix models $i$ and $j$.
For $k\in\{i,j\}$, let
\[
\ell_{ksm}
:=
\log p_k(y_{sm}|x_s),
\quad
r_{sm}
:=
\ell_{ism}-\ell_{jsm}.
\]
Thus, $r_{sm}=r_{ij}(x_s,y_{sm})$ in the notation of Appendix~\ref{app:conditional-kl-quadratic-approximations}.
For each $k\in\{i,j\}$, the globally centered coordinates are
\[
\begin{aligned}
\bar\ell_k^{(g)}
&:=
\frac{1}{NM}
\sum_{s=1}^N\sum_{m=1}^M\ell_{ksm},\\
\xi_{ksm}^{(g)}
&:=
\ell_{ksm}-\bar\ell_k^{(g)}.
\end{aligned}
\]
The corresponding coordinates obtained with per-prompt centering are
\[
\begin{aligned}
\bar\ell_{ks}^{(p)}
&:=
\frac1M\sum_{m=1}^M\ell_{ksm},\\
\xi_{ksm}^{(p)}
&:=
\ell_{ksm}-\bar\ell_{ks}^{(p)}.
\end{aligned}
\]
The empirical counterparts of (\ref{eq:kl-three-quadratic-quantities}) are as follows; the per-prompt estimator requires $M\geq2$:
\begin{align}
&\widehat{\mathcal{A}}_{\text{uncentered}}
:=
\frac{1}{NM}
\sum_{s=1}^N\sum_{m=1}^M r_{sm}^2,\nonumber\\
&\widehat{\mathcal{A}}_{\text{global}}
:=\nonumber\\[-2pt]
&\quad
\frac{1}{NM}
\sum_{s=1}^N\sum_{m=1}^M
\left(
\xi_{ism}^{(g)}-\xi_{jsm}^{(g)}
\right)^2,\nonumber\\
&\widehat{\mathcal{A}}_{\text{per-prompt}}
:=\nonumber\\[-2pt]
&\quad
\frac{1}{N(M-1)}
\sum_{s=1}^N\sum_{m=1}^M
\left(
\xi_{ism}^{(p)}-\xi_{jsm}^{(p)}
\right)^2.
\label{eq:kl-empirical-estimators}
\end{align}
The uncentered estimator is unbiased for
$\mathcal{A}_{\text{uncentered}}$ in finite samples.
The globally centered estimator is generally downward-biased in finite samples but is consistent for
$\mathcal{A}_{\text{global}}$ as $N\to\infty$ with fixed $M$.
The per-prompt estimator is the average of the usual unbiased within-prompt sample variances and is therefore unbiased for
$\mathcal{A}_{\text{per-prompt}}$ when $M\geq2$.

For the dataset $D$, where $M=1$, identifying $y_{s1}$ with $y_s$ gives $\xi_{is1}^{(g)}=\xi_{is}$, so $\widehat{\mathcal{A}}_{\text{global}}$ in (\ref{eq:kl-empirical-estimators}) is exactly the final expression in (\ref{eq:kl-global-coordinate-m1}).
Thus, (\ref{eq:kl-global-coordinate-m1}), equivalently (\ref{eq:kl-conditional-xi}), can be written as $2\KL(p_i,p_j)\approx\widehat{\mathcal{A}}_{\text{global}}$.
Moreover, $\widehat{\mathcal{A}}_{\text{uncentered}}=\|\bmell_i-\bmell_j\|^2/N$, the pairwise squared distance between the log-likelihood vectors visualized in our figures, divided by $N$.
In contrast, $\widehat{\mathcal{A}}_{\text{per-prompt}}$ is not available because a within-prompt variance cannot be estimated from one response.

\subsection{Finite-sample comparison of uncentered and global estimators}

We next compare the uncentered and globally centered estimators on the same finite sample and ask whether their difference affects the leading accuracy of the KL approximation.
Define the prompt-level and overall means of the observed log-likelihood ratios by
\[
\bar r_s
:=
\frac1M\sum_{m=1}^M r_{sm},
\quad
\bar r
:=
\frac1N\sum_{s=1}^N\bar r_s.
\]
The uncentered and globally centered quantities in (\ref{eq:kl-empirical-estimators}) satisfy the exact Pythagorean identity
\begin{align}
\widehat{\mathcal{A}}_{\text{uncentered}}
-
\widehat{\mathcal{A}}_{\text{global}}
=
\bar r^2.
\label{eq:kl-empirical-centering-gap}
\end{align}
To evaluate this gap, we account for the dependence among responses that share a prompt.
Conditional on $x_s$,
\[
\begin{aligned}
\bbE[\bar r_s|x_s]
&=
\mu_{ij}(x_s),\\
\Var(\bar r_s|x_s)
&=
\frac1M
\Var(r_{ij}(x_s,y)|x_s).
\end{aligned}
\]
The first identity and the law of total expectation give
$\bbE[\bar r]=\bbE[\bar r_s]
=\bbE\!\left[\bbE[\bar r_s|x_s]\right]=\mu_{ij}$.
Similarly, substituting the two conditional identities above into the law of total variance gives
$\Var(\bar r_s)=\Var_x(\mu_{ij}(x_s))
+\bbE_x[\Var(r_{ij}(x_s,y)|x_s)]/M
=V_{ij}+\mathcal{A}_{\text{per-prompt}}/M$.
Independence across $s$ then gives
\begin{align}
\Var(\bar r)
=
\frac{V_{ij}}{N}
+\frac{\mathcal{A}_{\text{per-prompt}}}{NM}.
\label{eq:kl-overall-mean-variance}
\end{align}
Taking expectations in (\ref{eq:kl-empirical-centering-gap}), using the unbiasedness of $\widehat{\mathcal{A}}_{\text{uncentered}}$, $\bbE[\bar r]=\mu_{ij}$, and the first identity in (\ref{eq:kl-centering-decomposition}), gives the exact bias
\[
\begin{aligned}
&\bbE[\widehat{\mathcal{A}}_{\text{global}}]
-\mathcal{A}_{\text{global}}
\\[-2pt]
&\quad=
\mathcal{A}_{\text{uncentered}}
-\bbE[\bar r^2]
-\mathcal{A}_{\text{global}}\\
&\quad=
\mathcal{A}_{\text{uncentered}}
-\mathcal{A}_{\text{global}}
-\mu_{ij}^2
-\Var(\bar r)\\
&\quad=
-\Var(\bar r)
=
-\frac{V_{ij}}{N}
-\frac{\mathcal{A}_{\text{per-prompt}}}{NM}.
\end{aligned}
\]
This explains the downward finite-sample bias noted above and shows that the bias vanishes as $N\to\infty$ with fixed $M$.

We next determine the asymptotic order of the empirical centering gap in (\ref{eq:kl-empirical-centering-gap}).
Under the local conditions of Appendix~\ref{app:conditional-kl-quadratic-approximations},
$V_{ij}=O(\lambda^4)$ and
$\mathcal{A}_{\text{per-prompt}}=O(\lambda^2)$.
Substituting these orders into (\ref{eq:kl-overall-mean-variance}) gives
\[
\Var(\bar r)
=
O\left(
\frac{\lambda^4}{N}
+
\frac{\lambda^2}{NM}
\right).
\]
Since $\bbE[\bar r]=\mu_{ij}$, the standard deviation of $\bar r$ gives the fluctuation order
\[
\bar r-\mu_{ij}
=
O_p\left(
\frac{\lambda^2}{\sqrt N}
+
\frac{\lambda}{\sqrt{NM}}
\right).
\]
Together with $\mu_{ij}=O(\lambda^2)$, this implies
$\bar r=O_p(\lambda^2+\lambda/\sqrt{NM})$.
Squaring this order and using the exact identity (\ref{eq:kl-empirical-centering-gap}) yields
\begin{align}
\widehat{\mathcal{A}}_{\text{uncentered}}
-
\widehat{\mathcal{A}}_{\text{global}}
=
O_p
\left(
\lambda^4+\frac{\lambda^2}{NM}
\right).
\label{eq:kl-empirical-gap-order}
\end{align}

\subsection{Sampling variance of the uncentered estimator}

We now derive the exact sampling variance of the uncentered estimator.
Let
\[
g(x,y)
:=
r_{ij}(x,y)^2,
\]
and define its between-prompt and average within-prompt variance components by
\[
\begin{aligned}
V_{\text{between}}
&:=
\Var_x
\left(
\bbE[g(x,y)|x]
\right),\\
V_{\text{within}}
&:=
\bbE_x
\left[
\Var(g(x,y)|x)
\right].
\end{aligned}
\]
Writing $\bar g_s=M^{-1}\sum_{m=1}^M g(x_s,y_{sm})$, we have $\widehat{\mathcal{A}}_{\text{uncentered}}=N^{-1}\sum_{s=1}^N\bar g_s$.
As in the derivation of (\ref{eq:kl-overall-mean-variance}), the law of total variance gives
\[
\begin{aligned}
\Var(\bar g_s)
&=
\Var_x
\left(
\bbE[\bar g_s|x_s]
\right)
+
\bbE_x
\left[
\Var(\bar g_s|x_s)
\right]\\
&=
V_{\text{between}}
+
\frac{V_{\text{within}}}{M}.
\end{aligned}
\]
Averaging over the independent prompt clusters then gives
\begin{align}
\Var
\left(
\widehat{\mathcal{A}}_{\text{uncentered}}
\right)
=
\frac{V_{\text{between}}}{N}
+\frac{V_{\text{within}}}{NM}.
\label{eq:kl-uncentered-allocation-variance}
\end{align}

\subsection{Leading accuracy of the uncentered and globally centered estimators}

We first derive the leading accuracy of the uncentered estimator and then use the empirical centering gap in (\ref{eq:kl-empirical-gap-order}) to obtain the corresponding result for the globally centered estimator.
Under the local conditions of Appendix~\ref{app:conditional-kl-quadratic-approximations}, $r_{ij}(x,y)=O_p(\lambda)$.
Together with suitable fourth-moment bounds, this gives
$V_{\text{between}}=O(\lambda^4)$ and
$V_{\text{within}}=O(\lambda^4)$.
Substituting these orders into (\ref{eq:kl-uncentered-allocation-variance}) yields
\[
\begin{aligned}
\Var
\left(
\widehat{\mathcal{A}}_{\text{uncentered}}
\right)
&=
O\left(
\frac{\lambda^4}{N}
+
\frac{\lambda^4}{NM}
\right)
\\
&=
O\left(
\frac{\lambda^4}{N}
\right).
\end{aligned}
\]
Because $\widehat{\mathcal{A}}_{\text{uncentered}}$ is unbiased for $\mathcal{A}_{\text{uncentered}}$, its standard deviation therefore gives
\[
\widehat{\mathcal{A}}_{\text{uncentered}}
-
\mathcal{A}_{\text{uncentered}}
=
O_p
\left(
\frac{\lambda^2}{\sqrt N}
\right).
\]
Together with (\ref{eq:kl-three-representations}), this gives
\begin{align}
\widehat{\mathcal{A}}_{\text{uncentered}}
-
2\KL(p_i,p_j)
=
O_p
\left(
\lambda^3+\frac{\lambda^2}{\sqrt N}
\right).
\label{eq:kl-uncentered-estimation-error}
\end{align}
When $\lambda\to0$ and $N\to\infty$ with fixed $M$, the terms in (\ref{eq:kl-empirical-gap-order}) are of lower order than those in (\ref{eq:kl-uncentered-estimation-error}).
Therefore,
\[
\widehat{\mathcal{A}}_{\text{global}}
-
2\KL(p_i,p_j)
=
O_p
\left(
\lambda^3+
\frac{\lambda^2}{\sqrt N}
\right),
\]
and the globally centered estimator has the same leading accuracy as the uncentered estimator.

\subsection{Optimization under a fixed response budget}

As a simple practical consequence of the variance formula in (\ref{eq:kl-uncentered-allocation-variance}), consider how to allocate a fixed response budget between the number $N$ of prompts and the number $M$ of responses per prompt.
We use an idealized setting in which the total number of responses
\[
T=NM
\]
is fixed, additional independent prompts are available, and prompt acquisition does not incur a separate cost in this budget.
Since $N=T/M$, (\ref{eq:kl-uncentered-allocation-variance}) becomes
\begin{align}
\Var
\left(
\widehat{\mathcal{A}}_{\text{uncentered}}
\right)
=
\frac{
M V_{\text{between}}
+V_{\text{within}}
}{T}.
\label{eq:kl-fixed-response-budget}
\end{align}
For fixed $T$, only the term $M V_{\text{between}}/T$ depends on $M$.
If $V_{\text{between}}>0$, this variance is minimized at $M=1$ among integer allocations satisfying $NM=T$.
Thus, under this simple response-budget model, the optimal allocation uses one response per prompt and includes as many independent prompts as the budget allows, rather than collecting multiple responses for fewer prompts.

For the globally centered estimator, (\ref{eq:kl-empirical-gap-order}) and $T=NM$ give
\[
\widehat{\mathcal{A}}_{\text{uncentered}}
-
\widehat{\mathcal{A}}_{\text{global}}
=
O_p\left(
\lambda^4+\frac{\lambda^2}{T}
\right).
\]
Here, the $\lambda^4$ term is lower order than the $O(\lambda^3)$ approximation error, while $\lambda^2/T=o(\lambda^2/\sqrt N)$ for fixed $M$ and is independent of the allocation at fixed $T$.
Thus, the centering gap does not change the leading-order allocation criterion in (\ref{eq:kl-fixed-response-budget}), and $M=1$ also minimizes the leading-order sampling variance of the globally centered estimator.

\subsection{Practical role of centering}

Although centering is not required for the second-order KL approximation, it changes which additive components are retained in the representation.
A common practical motivation for global centering is to remove a model-specific additive offset from log-likelihood vectors, as discussed in Appendix C of \citet{modelmap2025}.

To make these invariances explicit, consider transformed log-likelihoods
\[
\ell'_{ism}
:=
\ell_{ism}+a_i+b_{is},
\]
where $a_i$ is a model-specific constant and $b_{is}$ is an offset specific to model $i$ and prompt $x_s$, but common to all $M$ responses for that prompt.
Let
\[
\bar b_i
:=
\frac1N\sum_{s=1}^N b_{is}.
\]
The corresponding centered coordinates satisfy
\begin{align}
\xi_{ism}^{(g)\prime}
&=
\xi_{ism}^{(g)}
+b_{is}-\bar b_i,\nonumber\\
\xi_{ism}^{(p)\prime}
&=
\xi_{ism}^{(p)}.
\label{eq:centering-offsets}
\end{align}
Thus, uncentered coordinates retain both $a_i$ and $b_{is}$.
Global centering removes $a_i$, but generally retains the prompt-varying component $b_{is}-\bar b_i$.
Per-prompt centering removes the entire offset $a_i+b_{is}$ separately for each prompt.

Equation~(\ref{eq:kl-centering-decomposition}) quantifies exactly what the two centering operations remove:
\[
\begin{aligned}
\mathcal{A}_{\text{uncentered}}
-\mathcal{A}_{\text{global}}
&=
\mu_{ij}^2,\\
\mathcal{A}_{\text{global}}
-\mathcal{A}_{\text{per-prompt}}
&=
V_{ij}.
\end{aligned}
\]
Thus, global centering removes the squared overall mean log-likelihood difference $\mu_{ij}^2$ but retains the across-prompt variation $V_{ij}$ in the conditional mean $\mu_{ij}(x)$.
Per-prompt centering removes both components.

Under the local conditions of Appendix~\ref{app:conditional-kl-quadratic-approximations}, $\mu_{ij}^2$ and $V_{ij}$ are both $O(\lambda^4)$, so removing them does not affect the second-order KL approximation.
Away from the local regime, however, these terms can change pairwise model distances.
Our visualizations use uncentered vectors and therefore retain both components.
Global or per-prompt centering is appropriate when invariance to the corresponding additive offsets is desired; in particular, global centering can remove a meaningful overall mean difference as well as a nuisance constant offset.
\section{Comparison of MI Estimates Using Model and Reference Marginals}
\label{app:marginal-proxy}

Section~\ref{sec:pmi-vector} distinguishes two response marginals.
The reference marginal in (\ref{eq:reference-marginal}) is induced by the
prompt-response distribution $p_i(y|x)p_0(x)$ and is required for the PMI
interpretation of (\ref{eq:diff-is}). The model marginal in
(\ref{eq:model-marginal}) is induced by the model's own prompt distribution.
In practice, we do not directly evaluate either expectation. We use
empty-prompt evaluation to estimate the model marginal in the main experiments,
and use shuffled-prompt Monte Carlo estimation to estimate the reference
marginal in this comparison. Because the two marginals average over
different prompt distributions, we do not expect their estimates to agree
numerically. Instead, we examine whether the empty-prompt estimate provides a
useful practical substitute for the reference marginal when comparing the
model-level MI estimates in (\ref{eq:mi-bar-delta-ell}) and
(\ref{eq:mi-eta}).

\begin{figure*}[!t]
\centering
\includegraphics[width=\linewidth]{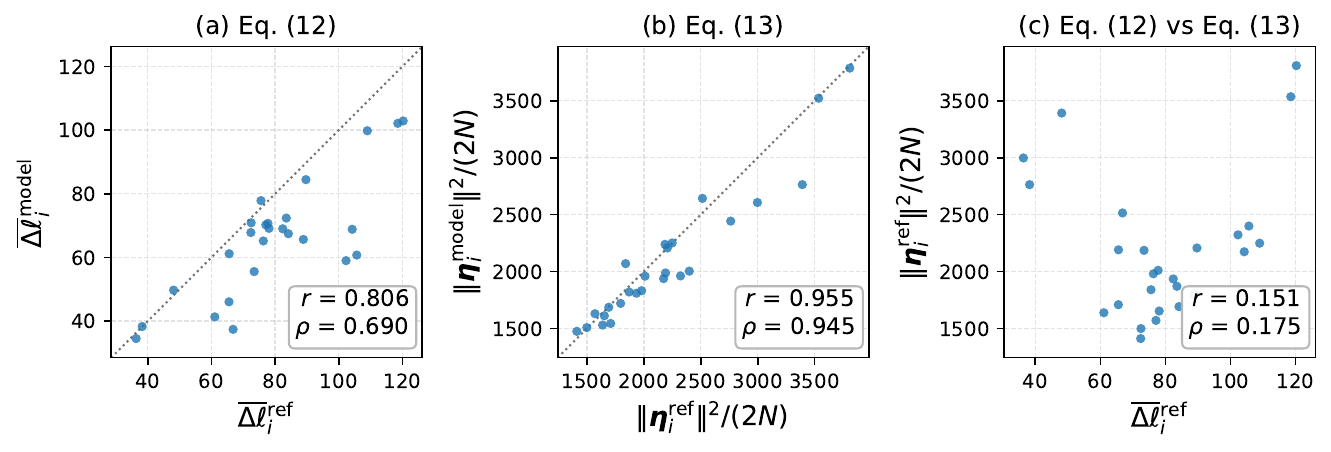}
\caption{Comparison of MI estimates using the model and reference marginals.
We use $K'=26$ models, $C=10$ shuffled-prompt permutations, and $N=9{,}691$
prompt-response pairs; $r$ and $\rho$ denote Pearson and Spearman correlations
across the 26 models. In (a) and (b), the horizontal axis uses shuffled-prompt
Monte Carlo estimation of the reference marginal, whereas the vertical axis
uses the empty-prompt estimate of the model marginal; the dotted line is the
diagonal. Panel (a) shows the sample-mean estimate of MI in
(\ref{eq:mi-bar-delta-ell}), and panel (b) shows the squared-norm estimate of MI in
(\ref{eq:mi-eta}). Panel (c) compares the two estimates when both use the
shuffled-prompt estimate of the reference marginal.}
\label{fig:marginal-proxy}
\end{figure*}

\subsection{Marginal Estimators}

Let $x_{\emptyset}$ denote the input obtained by leaving the user-input field
empty under the same chat template as in the main experiments. We write the
empty-prompt estimate of the model marginal as
\begin{align*}
\widehat p_i^{\mathrm{model}}(y_s)
&:=p_i(y_s|x_{\emptyset}).
\end{align*}
The hat emphasizes that we do not compute the expectation over $p_i(x)$ in
(\ref{eq:model-marginal}).

Let $\sigma_1,\ldots,\sigma_C$ be independent uniformly random permutations of
$\{1,\ldots,N\}$ satisfying $\sigma_c(s)\ne s$ for every $s$. We define the shuffled-prompt
Monte Carlo estimate of the reference marginal by
\begin{align*}
\widehat p_i^{\mathrm{ref}}(y_s)
&:=
\frac{1}{C}\sum_{c=1}^{C}
p_i(y_s|x_{\sigma_c(s)}).
\end{align*}
This estimator averages probabilities, rather than log probabilities, over
$C$ prompts paired with the fixed response $y_s$. Conditional on the observed
prompt set, its expectation over these permutations is the leave-one-out average
\begin{align*}
\bbE_{\sigma}
\left[\widehat p_i^{\mathrm{ref}}(y_s)\right]
&=
\frac{1}{N-1}\sum_{t\ne s}p_i(y_s|x_t),
\end{align*}
which approximates (\ref{eq:reference-marginal}) under the empirical reference
prompt distribution.

\begin{samepage}
To distinguish the two treatments of the marginal, define
\begin{gather*}
\Delta\ell_{is}^{\mathrm{model}}
:=\log p_i(y_s|x_s)-\log\widehat p_i^{\mathrm{model}}(y_s),
\\
\Delta\ell_{is}^{\mathrm{ref}}
:=\log p_i(y_s|x_s)-\log\widehat p_i^{\mathrm{ref}}(y_s).
\end{gather*}
\end{samepage}
We use the same superscripts for the derived means and centered vectors, denoting
them by $\overline{\Delta\ell}_i^{\mathrm{model}}$,
$\overline{\Delta\ell}_i^{\mathrm{ref}}$,
$\bmeta_i^{\mathrm{model}}$, and $\bmeta_i^{\mathrm{ref}}$.

\subsection{Experimental Setup}

We use the same $C=10$ shuffled-prompt sets for all models. Starting with
10,000 prompt-response pairs from the Tulu3 text set, we exclude 309 pairs
containing at least one log-likelihood flagged as invalid during likelihood
computation, leaving $N=9{,}691$.
For the shuffled-prompt estimate, we average in probability space before taking
the logarithm. We then separately clip each of the conditional, empty-prompt,
and shuffled-prompt log-likelihood matrices at its 2nd percentile.

\paragraph{Language models.}
In this experiment we use the following $K'=26$ language models:
\begin{itemize}[nosep, leftmargin=0.7em]
\item \url{01-ai/Yi-1.5-9B-Chat}
\item \url{01-ai/Yi-1.5-9B-Chat-16K}
\item \url{BAAI/Infinity-Instruct-3M-0625-Qwen2-7B}
\item \url{BAAI/Infinity-Instruct-3M-0625-Yi-1.5-9B}
\item \url{Qwen/Qwen2-7B}
\item \url{Qwen/Qwen2-7B-Instruct}
\item \url{allenai/Llama-3.1-Tulu-3-8B}
\item \url{allenai/Llama-3.1-Tulu-3-8B-DPO}
\item \url{allenai/Llama-3.1-Tulu-3-8B-SFT}
\item \url{meta-llama/Llama-3.1-8B-Instruct}
\item \url{meta-llama/Meta-Llama-3-8B-Instruct}
\item \url{princeton-nlp/Llama-3-Base-8B-SFT}
\item \url{princeton-nlp/Llama-3-Base-8B-SFT-CPO}
\item \url{princeton-nlp/Llama-3-Base-8B-SFT-DPO}
\item \url{princeton-nlp/Llama-3-Base-8B-SFT-KTO}
\item \url{princeton-nlp/Llama-3-Base-8B-SFT-ORPO}
\item \url{princeton-nlp/Llama-3-Base-8B-SFT-RDPO}
\item \url{princeton-nlp/Llama-3-Base-8B-SFT-RRHF}
\item \url{princeton-nlp/Llama-3-Base-8B-SFT-SLiC-HF}
\item \url{princeton-nlp/Llama-3-Instruct-8B-CPO}
\item \url{princeton-nlp/Llama-3-Instruct-8B-DPO}
\item \url{princeton-nlp/Llama-3-Instruct-8B-KTO}
\item \url{princeton-nlp/Llama-3-Instruct-8B-ORPO}
\item \url{princeton-nlp/Llama-3-Instruct-8B-RDPO}
\item \url{princeton-nlp/Llama-3-Instruct-8B-RRHF}
\item \url{princeton-nlp/Llama-3-Instruct-8B-SLiC-HF}
\end{itemize}

\subsection{Results}

Figure~\ref{fig:marginal-proxy} reports the results.
For the sample-mean estimate of MI in (\ref{eq:mi-bar-delta-ell}), the estimates based
on the empty-prompt estimate of the model marginal and the shuffled-prompt
estimate of the reference marginal have Pearson correlation $r=0.806$ and
Spearman correlation $\rho=0.690$, as shown in
Figure~\ref{fig:marginal-proxy}(a). The two
marginals need not yield the same numerical values; these positive correlations
instead suggest that the empty-prompt evaluation retains useful information for
model-level comparison.

Figure~\ref{fig:marginal-proxy}(b) shows stronger corresponding correlations
for the squared-norm estimate of MI in (\ref{eq:mi-eta}): $r=0.955$ and
$\rho=0.945$. On this restricted model set and text set,
the result supports using the empty-prompt estimate as a practical substitute
for the reference marginal in the analyses reported in this paper.

Finally, Figure~\ref{fig:marginal-proxy}(c) compares the two MI estimates while
using the same shuffled-prompt estimate of the reference marginal for both.
Their correlation is positive but weak ($r=0.151$ and $\rho=0.175$). This weak
correlation is consistent with the status of (\ref{eq:mi-eta}) as only a rough
proxy for MI:
even when the same reference-marginal estimate is used, (\ref{eq:mi-eta})
additionally relies on the local
variance approximation whose closeness assumptions are restrictive when a
model's responses depend strongly on its prompts.

\section{Details of Experiments in Section~\ref{sec:conditional-model-map}}
\label{app:details-conditional-model-map}

See Appendix~\ref{sec:experimental-settings} for the text sets, the language models, and the computation of the log-likelihood.

\paragraph{Conditional model map.}
For the visualization of the conditional model map in Figure~\ref{fig:intro2} and the prompt-shift visualizations in Figure~\ref{fig:intro}, corresponding to Section~\ref{sec:conditional-model-attributes}, Section~\ref{sec:prompt-shift}, and Section~\ref{sec:addcomp-prompt-shift}, we use $K=552$ language models and $N=10{,}000$ prompt-response pairs sampled from Tulu3. For the conditional log-likelihood matrix $\bm{L}=[\bm{\ell}_1,\ldots,\bm{\ell}_K]^{\top}\in\mathbb{R}^{K\times N}$, entries in the bottom 2\% are clipped to the 2\%-percentile value.

\paragraph{Prompts used for prompt-shift experiments.}
In the experiments in Section~\ref{sec:prompt-shift} and Section~\ref{sec:addcomp-prompt-shift}, we use the following three prompt variants in addition to the original prompt $x$:
\begin{itemize}
  \item $T_{\mathrm{cot}}(x)$: The suffix ``\textit{Let's think step by step.}'' is appended to the original prompt, separated by a newline ($x\backslash n\,\textit{Let's think step by step.}$).
  \item $T_{\mathrm{rep}}(x)$: The original prompt is repeated twice, separated by two newlines ($x\backslash n\backslash n\,x$).
  \item $T_{\mathrm{rep+cot}}(x)$: The CoT suffix is further appended to the repeated prompt ($x\backslash n\backslash n\,x\backslash n\,\textit{Let's think step by step.}$).
\end{itemize}

\paragraph{PMI model map and mutual information.}
For the visualizations of the PMI model map and the conditional model map in Section~\ref{sec:pmi-map}, we use $K=481$ language models. We use two text sets: $N=10{,}000$ prompt-response pairs sampled from Tulu3 and $N=10{,}000$ prompt-response pairs sampled from Infinity-Instruct. Based on these two text sets, we compute $\bm{L}_{\mathrm{Tulu}}\in\mathbb{R}^{K\times N}$ and $\bm{L}_{\mathrm{Infinity}}\in\mathbb{R}^{K\times N}$, respectively. For each matrix, entries in the bottom 2\% are clipped to the 2\%-percentile value.

\begin{table*}[t]
\centering
\begin{adjustbox}{width=\linewidth}
\begin{tabular}{lrrccccc}
\toprule
 & & & \multicolumn{2}{c}{Relative error in $\bm{\ell}$} & \multicolumn{2}{c}{Relative error in $\bm{\xi}$} & \\
\cmidrule(lr){4-5}\cmidrule(lr){6-7}
Text set & $K$ & $N$ & Median & Mean & Median & Mean & Pearson $r$ of $\delta$ \\
\midrule
Tulu3  & $552$ & $10{,}000$ & 0.062 [0.058, 0.066] & 0.081 [0.075, 0.088] & 0.074 [0.070, 0.080] & 0.095 [0.088, 0.102] & 0.987 [0.959, 0.995] \\
Tulu3             & $100$ & $5{,}000$  & 0.064 [0.054, 0.070] & 0.070 [0.062, 0.079] & 0.070 [0.064, 0.081] & 0.083 [0.074, 0.094] & 0.984 [0.970, 0.992] \\
Infinity-Instruct & $100$ & $5{,}000$  & 0.083 [0.073, 0.102] & 0.097 [0.086, 0.109] & 0.100 [0.090, 0.109] & 0.116 [0.104, 0.129] & 0.986 [0.974, 0.993] \\
UltraChat         & $100$ & $5{,}000$  & 0.061 [0.048, 0.081] & 0.084 [0.072, 0.098] & 0.087 [0.074, 0.111] & 0.109 [0.095, 0.122] & 0.975 [0.964, 0.990] \\
\bottomrule
\end{tabular}
\end{adjustbox}
\caption{Additive compositionality of prompt shifts on the three text sets. The first four numeric columns report the relative prediction error of Eq.~(\ref{eq:v-additive-compositionality}) in the $\bm{\ell}$ and $\bm{\xi}$ spaces; the last column reports the Pearson correlation between $\delta_{\mathrm{rep+cot}}$ and $\delta_{\mathrm{cot}}+\delta_{\mathrm{rep}}$. Brackets denote bootstrap 95\% confidence intervals.}
\label{tab:app-additivity-datasets}
\end{table*}

\begin{table*}[t]
\centering
\begin{adjustbox}{width=\linewidth}
\begin{tabular}{llccccc@{\hspace{1.5em}}c}
\toprule
 & Text set & ARC-Challenge & ARC-Easy & OpenBookQA & PIQA & TruthfulQA & Mean \\
\midrule
\multirow{4}{*}{Pearson $r$}
 & Tulu3 ($K{=}485$) & 0.829 [0.783, 0.867] & 0.841 [0.803, 0.872] & 0.586 [0.363, 0.807] & 0.906 [0.881, 0.931] & 0.842 [0.790, 0.890] & 0.801 \\
 & Tulu3 ($K{=}100$) & 0.892 [0.849, 0.924] & 0.784 [0.700, 0.850] & 0.805 [0.724, 0.864] & 0.910 [0.867, 0.941] & 0.766 [0.647, 0.861] & 0.831 \\
 & Infinity-Instruct & 0.905 [0.862, 0.937] & 0.739 [0.638, 0.830] & 0.766 [0.644, 0.866] & 0.879 [0.805, 0.932] & 0.832 [0.733, 0.893] & 0.824 \\
 & UltraChat         & 0.775 [0.687, 0.844] & 0.588 [0.443, 0.704] & 0.779 [0.700, 0.843] & 0.868 [0.741, 0.940] & 0.822 [0.740, 0.885] & 0.766 \\
\midrule
\multirow{4}{*}{Spearman $\rho$}
 & Tulu3 ($K{=}485$) & 0.821 [0.765, 0.869] & 0.824 [0.778, 0.863] & 0.797 [0.728, 0.854] & 0.945 [0.926, 0.960] & 0.869 [0.820, 0.906] & 0.851 \\
 & Tulu3 ($K{=}100$) & 0.880 [0.802, 0.927] & 0.765 [0.643, 0.845] & 0.805 [0.696, 0.873] & 0.899 [0.825, 0.938] & 0.796 [0.662, 0.887] & 0.829 \\
 & Infinity-Instruct & 0.904 [0.836, 0.942] & 0.739 [0.605, 0.833] & 0.735 [0.571, 0.847] & 0.851 [0.739, 0.922] & 0.763 [0.621, 0.866] & 0.798 \\
 & UltraChat         & 0.799 [0.688, 0.869] & 0.638 [0.475, 0.761] & 0.758 [0.639, 0.840] & 0.850 [0.709, 0.941] & 0.841 [0.740, 0.907] & 0.777 \\
\bottomrule
\end{tabular}
\end{adjustbox}
\caption{Correlation between predicted and true task scores on the test set for the three text sets, corresponding to Table~\ref{tab:score-pred}. Tulu3 ($K{=}100$), Infinity-Instruct, and UltraChat use the matched setting ($K{=}100$ models, $N{=}5{,}000$ texts). Brackets denote bootstrap 95\% confidence intervals. The mean value across five tasks is shown in the right column.}
\label{tab:app-score-pred-datasets}
\end{table*}

\begin{table*}[t]
\centering
\begin{adjustbox}{width=\linewidth}
\begin{tabular}{llccccc@{\hspace{1.5em}}c}
\toprule
 & Text set & ARC-Challenge & ARC-Easy & OpenBookQA & PIQA & TruthfulQA & Mean\\
\midrule
\multirow{4}{*}{$\Delta$\textsc{cot}}
 & Tulu3 ($K{=}485$)   & 0.694 [0.541, 0.827] & 0.728 [0.625, 0.817] & 0.595 [0.459, 0.701] & 0.678 [0.529, 0.794] & 0.486 [0.327, 0.648] & 0.636 \\
 & Tulu3 ($K{=}100$)   & 0.674 [0.407, 0.845] & 0.615 [0.269, 0.872] & 0.491 [$-0.148$, 0.839] & 0.140 [$-0.283$, 0.747] & 0.262 [$-0.255$, 0.677] & 0.436 \\
 & Infinity-Instruct   & 0.551 [0.171, 0.805] & 0.679 [0.457, 0.857] & 0.636 [0.233, 0.847] & 0.627 [0.372, 0.850] & 0.481 [0.112, 0.717] & 0.595 \\
 & UltraChat           & 0.452 [0.061, 0.732] & 0.331 [$-0.221$, 0.800] & 0.567 [0.070, 0.839] & 0.398 [$-0.056$, 0.772] & 0.301 [$-0.255$, 0.667] & 0.410 \\
\midrule
\multirow{4}{*}{$\Delta$\textsc{repeat}}
 & Tulu3 ($K{=}485$)   & 0.678 [0.567, 0.773] & 0.359 [0.155, 0.540] & 0.250 [$-0.021$, 0.500] & 0.408 [0.135, 0.610] & 0.557 [0.405, 0.691] & 0.450 \\
 & Tulu3 ($K{=}100$)   & 0.635 [0.213, 0.836] & 0.147 [$-0.210$, 0.511] & 0.254 [$-0.361$, 0.748] & 0.337 [$-0.060$, 0.760] & 0.003 [$-0.443$, 0.595] & 0.275 \\
 & Infinity-Instruct   & 0.369 [$-0.091$, 0.690] & 0.385 [$-0.193$, 0.749] & $-0.009$ [$-0.565$, 0.458] & $-0.236$ [$-0.610$, 0.520] & 0.513 [0.205, 0.786] & 0.204 \\
 & UltraChat           & 0.502 [0.205, 0.745] & 0.388 [$-0.044$, 0.694] & 0.144 [$-0.376$, 0.692] & 0.001 [$-0.477$, 0.584] & 0.473 [0.227, 0.765] & 0.302 \\
\midrule
\multirow{4}{*}{$\Delta$\textsc{repeat+cot}}
 & Tulu3 ($K{=}485$)   & 0.588 [0.366, 0.778] & 0.603 [0.448, 0.735] & 0.501 [0.323, 0.662] & 0.518 [0.290, 0.702] & 0.721 [0.624, 0.799] & 0.586 \\
 & Tulu3 ($K{=}100$)   & 0.601 [0.298, 0.813] & 0.578 [0.103, 0.866] & 0.434 [$-0.094$, 0.824] & 0.518 [0.080, 0.894] & 0.164 [$-0.313$, 0.756] & 0.459 \\
 & Infinity-Instruct   & 0.440 [0.118, 0.724] & 0.704 [0.429, 0.898] & 0.409 [$-0.115$, 0.871] & 0.499 [0.111, 0.822] & 0.394 [$-0.002$, 0.760] & 0.489 \\
 & UltraChat           & 0.584 [0.209, 0.808] & 0.515 [0.083, 0.776] & 0.366 [$-0.146$, 0.763] & 0.669 [0.320, 0.838] & 0.228 [$-0.206$, 0.638] & 0.472 \\
\bottomrule
\end{tabular}
\end{adjustbox}
\caption{Correlation (Pearson $r$) between predicted and true score changes induced by each prompt transformation on the test set for the three text sets, corresponding to Table~\ref{tab:delta-score-pred}. Tulu3 ($K{=}100$), Infinity-Instruct, and UltraChat use the matched setting ($K{=}100$ models, $N{=}5{,}000$ texts). Brackets denote bootstrap 95\% confidence intervals. The mean value across five tasks is shown in the right column.}
\label{tab:app-delta-score-pred-datasets}
\end{table*}

\begin{table*}[t]
\centering
\begin{adjustbox}{width=0.9\linewidth}
\begin{tabular}{lcccc}
\toprule
 & \multicolumn{2}{c}{Pearson $r$} & \multicolumn{2}{c}{Spearman $\rho$} \\
\cmidrule(lr){2-3}\cmidrule(lr){4-5}
Task & Conditional & Unconditional & Conditional & Unconditional \\
\midrule
ARC-Challenge & 0.914 [0.864, 0.950] & 0.914 [0.868, 0.950] & 0.895 [0.821, 0.939] & 0.897 [0.826, 0.938] \\
ARC-Easy      & 0.864 [0.791, 0.914] & 0.839 [0.757, 0.901] & 0.861 [0.773, 0.917] & 0.856 [0.779, 0.910] \\
OpenBookQA    & 0.879 [0.819, 0.929] & 0.862 [0.733, 0.943] & 0.858 [0.769, 0.914] & 0.848 [0.746, 0.913] \\
PIQA          & 0.945 [0.904, 0.972] & 0.920 [0.867, 0.965] & 0.942 [0.901, 0.967] & 0.943 [0.899, 0.962] \\
TruthfulQA    & 0.902 [0.830, 0.950] & 0.884 [0.808, 0.935] & 0.901 [0.828, 0.942] & 0.873 [0.776, 0.935] \\
\midrule
Mean & 0.901 & 0.884 & 0.891 & 0.883 \\
\bottomrule
\end{tabular}
\end{adjustbox}
\caption{Correlation between predicted and true task scores for conditional ($\log p(y|x)$) and unconditional ($\log p(y)$) log-likelihood vectors, on the $K{=}438$ models for which both representations are available. Brackets denote bootstrap 95\% confidence intervals. The mean value across five tasks is shown in the bottom row.}
\label{tab:app-unconditional}
\end{table*}

\section{Results on Additional Text Sets}
\label{app:additional-text-sets}

Most analyses in the main text are based on the Tulu3 text set.
To examine the dependence of our findings on this choice, we conduct the main analyses on two additional text sets. The first set contains $N=5{,}000$ prompt-response pairs extracted from Infinity-Instruct and the second set contains $N=5{,}000$ pairs extracted from UltraChat~\citep{ding-etal-2023-enhancing}\footnote{\url{https://huggingface.co/datasets/HuggingFaceH4/ultrachat_200k}}.
For these experiments, we use $K=100$ models randomly selected from the $552$ models in Section~\ref{sec:conditional-model-map}, and compute the conditional log-likelihood vectors under the four prompt settings for each text set.
As a reference, we also report the results on Tulu3.
Confidence intervals are computed in the same way as in Section~\ref{sec:score-prediction}, with $20$ held-out test models in the $K=100$ settings.

\paragraph{Additive compositionality of prompt shifts.}
Table~\ref{tab:app-additivity-datasets} shows the relative prediction errors of the additive compositionality in Section~\ref{sec:addcomp-prompt-shift} and the correlation between $\delta_{\mathrm{rep+cot}}$ and $\delta_{\mathrm{cot}}+\delta_{\mathrm{rep}}$.
Both are at comparable levels across the three text sets.

\paragraph{Task score prediction.}
Table~\ref{tab:app-score-pred-datasets} shows the correlation between predicted and true task scores, corresponding to Table~\ref{tab:score-pred}. The mean value of the Pearson $r$ is within a range of 0.77 to 0.83 across the three text sets, indicating that the prediction performance is comparable regardless of the choice of text set.

\paragraph{Predicting score changes via shift vectors.}
Table~\ref{tab:app-delta-score-pred-datasets} shows the predicted score changes corresponding to Table~\ref{tab:delta-score-pred}.
With $K{=}100$ models, the mean value of the Pearson $r$ is within a range of 0.41 to 0.60 for $\Delta$\textsc{cot}, 0.20 to 0.30 for $\Delta$\textsc{repeat}, and 0.46 to 0.49 for $\Delta$\textsc{repeat+cot}, indicating that the overall trends are comparable regardless of the choice of text set.

\section{Comparison with the Unconditional Baseline}
\label{app:unconditional-baseline}

We compare conditional log-likelihood vectors and unconditional log-likelihood vectors~\citep{modelmap2025} as features for task score prediction.
We use the Tulu3 text set ($N=10{,}000$) and the $K=438$ models for which the conditional log-likelihoods, the unconditional log-likelihoods, and the task scores are available.
Unlike Section~\ref{sec:score-prediction}, each representation is used as a single feature matrix without concatenating prompt conditions, and the conditional log-likelihoods are computed under the \textsc{base} setting; the values are therefore not directly comparable to those in Table~\ref{tab:score-pred}.
We train RidgeCV models with an 8:2 model-wise split, and confidence intervals are 95\% percentile bootstrap intervals obtained by resampling the $88$ held-out test models with replacement ($B{=}2{,}000$), keeping the fitted regression models fixed. Table~\ref{tab:app-unconditional} shows that the two representations achieve comparable prediction performance on all five tasks. On average, the conditional representation is higher by just 1.7 points in Pearson $r$ and by 0.8 points in Spearman $\rho$.
The gain is small, and we regard the main merit of the conditional representation as enabling analyses that the unconditional representation cannot support, such as the prompt shifts in Section~\ref{sec:prompt-shift} and the prediction of prompt-induced score changes in Section~\ref{sec:score-prediction}.

\section{Details of Experiments in Section~\ref{sec:empirical-validation}}
\label{app:details-empirical-validation}

See Appendix~\ref{sec:experimental-settings} for the text sets, the language models, and the computation of the log-likelihood.

\paragraph{Language models.}
In the experiment in Section~\ref{sec:empirical-validation}, we use the following $K'=8$ instruction-tuned language models:
\begin{itemize}[nosep]
\item \url{Qwen/Qwen2-7B-Instruct}
\item \url{Qwen/Qwen2.5-7B-Instruct}
\item \url{google/gemma-2-9b-it}
\item \url{meta-llama/Llama-3.1-8B-Instruct}
\item \url{meta-llama/Llama-3.2-3B-Instruct}
\item \url{meta-llama/Meta-Llama-3-8B-Instruct}
\item \url{microsoft/Phi-3-mini-128k-instruct}
\item \url{mistralai/Mistral-7B-Instruct-v0.3}
\end{itemize}

\paragraph{Prompts.}
We sample $N=10{,}000$ prompts $x_1,\dots,x_N$ from \texttt{tulu-3-sft-mixture}~\cite{lambert2025tulu} and use only the prompts for generation.

\paragraph{Generation settings.}
In both Section~\ref{sec:monte-carlo-kl} and Section~\ref{sec:semantic-distance}, responses are generated using vLLM~\cite{kwon2023efficient} with bfloat16 precision. The sampling hyperparameters are summarized in Table~\ref{tab:gen_settings}.

\begin{table}[ht]
  \centering
  \begin{tabular}{lr}
    \toprule
    Parameter & Value \\
    \midrule
    Temperature (sampling)     & 1.0 \\
    Top-$p$ (nucleus sampling) & 0.95 \\
    Top-$k$                    & 50 \\
    Max.\ response tokens      & 768 \\
    \bottomrule
  \end{tabular}
  \caption{Hyperparameters for text generation.}
  \label{tab:gen_settings}
\end{table}

\paragraph{Monte Carlo estimate of KL divergence.}
For the experiment in Section~\ref{sec:monte-carlo-kl}, we generate one response $y_s^i \sim p_i(\cdot|x_s)$ for each prompt $x_s$ and each model $p_i$. For each ordered model pair $(p_i,p_j)$, we compute the log-likelihood $\log p_j(y_s^i|x_s)$ of texts generated by model $p_i$, and estimate $\mathrm{KL}(p_i,p_j)$ by the Monte Carlo approximation corresponding to Eq.~(\ref{eq:kl-conditional}):
\begin{align*}
\mathrm{KL}(p_i, p_j)
&=
\mathbb{E}_{(x,y)\sim p_i(y|x)p_0(x)}
\left[
\log \frac{p_i(y|x)}{p_j(y|x)}
\right] \\
&\approx
\frac{1}{N}\sum_{s=1}^{N}
\log \frac{p_i(y_s^i|x_s)}{p_j(y_s^i|x_s)}.
\end{align*}
The estimate based on log-likelihood vectors in (\ref{eq:kl-conditional-xi}) is symmetric in the two models. To compare quantities of the same kind, we symmetrize the Monte Carlo estimate as
\[
\frac{1}{2}\bigl(\mathrm{KL}(p_i,p_j)+\mathrm{KL}(p_j,p_i)\bigr),
\]
where both directions are estimated by the corresponding Monte Carlo approximations above, and we plot this symmetrized quantity in Figure~\ref{fig:kl_vs_pseudo_kl_scatter}.

\paragraph{KL divergence based on log-likelihood vectors.}
For the comparison in Section~\ref{sec:monte-carlo-kl}, each of the $K'$ models generates one response for each of the $N$ prompts, yielding $K'N$ prompt-response pairs in total. For each model $p_j$, we compute the conditional log-likelihood $\log p_j(y_s^i|x_s)$ for all these pairs and use the resulting values as components of a log-likelihood vector. We center each vector globally as in Section~\ref{sec:log-likelihood-vector}, and estimate pairwise KL divergence from the squared Euclidean distance between the centered vectors following (\ref{eq:kl-conditional-xi}).

\paragraph{Generated texts for semantic distance.}
For the experiment in Section~\ref{sec:semantic-distance}, we generate $M=10$ responses
\[
y_{s1}^i,\dots,y_{sM}^i \sim p_i(\cdot|x_s),
\]
for each prompt $x_s$ and each model $p_i$.

\paragraph{Text embedding model.}
In Section~\ref{sec:semantic-distance}, we compute embeddings of generated texts using the Sentence-Transformer model \texttt{all-MiniLM-L6-v2}~\cite{reimers-gurevych-2019-sentence}, which produces 384-dimensional dense vectors.

\paragraph{Semantic distance.}
For the comparison in Section~\ref{sec:semantic-distance}, we compute a semantic distance based on pairwise cosine similarities between generated-response embeddings. Let $\mathrm{emb}(y)\in\mathbb{R}^d$ denote the normalized embedding of text $y$. For each prompt $x_s$ and model $p_i$, we define the mean embedding vector
\[
\bar{v}_s^i
=
\frac{1}{M}
\sum_{m=1}^{M}
\mathrm{emb}(y_{sm}^i).
\]
We then define the semantic distance by
\[
\mathrm{SemDist}(p_i,p_j)
:=
\frac{1}{N}
\sum_{s=1}^{N}
\left(
1-\langle \bar{v}_s^i,\bar{v}_s^j\rangle
\right),
\]
where $\langle \cdot,\cdot\rangle$ denotes the inner product.

To aid interpretation, write $e_{sm}^i:=\mathrm{emb}(y_{sm}^i)$. Expanding the mean embeddings gives
\[
\begin{aligned}
&\mathrm{SemDist}(p_i,p_j)
\\
&=
1-\frac{1}{NM^2}
\sum_{s=1}^{N}
\left\langle
\sum_{m=1}^{M} e_{sm}^i,
\sum_{m'=1}^{M} e_{sm'}^j
\right\rangle
\\
&=
1-\frac{1}{NM^2}
\sum_{s=1}^{N}
\sum_{m,m'=1}^{M}
\left\langle e_{sm}^i,e_{sm'}^j\right\rangle
\\
&=
1-\frac{1}{NM^2}
\sum_{s=1}^{N}
\sum_{m,m'=1}^{M}
\cos\!\left(e_{sm}^i,e_{sm'}^j\right).
\end{aligned}
\]
The first equality substitutes the definitions of the mean embeddings, the second uses bilinearity of the inner product, and the third uses the normalization of each embedding. Thus, the semantic distance is one minus the average, over prompts, of the mean pairwise cosine similarity across the $M^2$ response pairs from the two models. This average pairwise cosine similarity is computed as in \citet{jiang2025artificial}; taking its complement makes larger values indicate greater model dissimilarity.

\paragraph{Bootstrap confidence intervals.}
Bootstrap resampling is performed over the $N$ prompts. For KL divergence, we resample prompt-level units and recompute pairwise KL estimates. For semantic distance, we likewise resample prompts and recompute $\mathrm{SemDist}(p_i,p_j)$ using the corresponding mean embeddings for each resampled prompt.

\section{Results on Additional Language Models}
\label{app:24models}

In Section~\ref{sec:monte-carlo-kl}, we compared the log-likelihood-vector-based KL estimate with a Monte Carlo estimate using $K'=8$ models. Here, we repeat the comparison with a broader set of $K'=24$ models.

\paragraph{Language models.}
We use the following $K'=24$ instruction-tuned language models of various sizes and architectures, selected from the set of $K=552$ models described in Appendix~\ref{sec:experimental-settings}:
\begin{itemize}[nosep]
\item \url{01-ai/Yi-1.5-9B-Chat-16K}
\item \url{BramVanroy/fietje-2-instruct}
\item \url{IlyaGusev/gemma-2-2b-it-abliterated}
\item \url{MaziyarPanahi/calme-3.2-instruct-3b}
\item \url{MaziyarPanahi/calme-3.3-instruct-3b}
\item \url{Qwen/Qwen1.5-0.5B-Chat}
\item \url{Qwen/Qwen2-0.5B-Instruct}
\item \url{Qwen/Qwen2-7B-Instruct}
\item \url{Qwen/Qwen2.5-7B-Instruct}
\item \url{Qwen/Qwen2.5-7B-Instruct-1M}
\item \url{Qwen/Qwen2.5-Coder-14B-Instruct}
\item \url{google/gemma-2-9b-it}
\item \url{google/gemma-2b-it}
\item \url{grimjim/Llama-3-Instruct-8B-SPPO-Iter3-SimPO-merge}
\item \url{lmsys/vicuna-7b-v1.5}
\item \url{lordjia/Llama-3-Cantonese-8B-Instruct}
\item \url{meta-llama/Llama-3.1-8B-Instruct}
\item \url{meta-llama/Llama-3.2-3B-Instruct}
\item \url{meta-llama/Meta-Llama-3-8B-Instruct}
\item \url{microsoft/Phi-3.5-mini-instruct}
\item \url{mistralai/Mistral-7B-Instruct-v0.3}
\item \url{princeton-nlp/Mistral-7B-Instruct-CPO}
\item \url{speakleash/Bielik-11B-v2.3-Instruct}
\item \url{v000000/Qwen2.5-14B-Gutenberg-Instruct-Slerpeno}
\end{itemize}

\paragraph{Prompts.}
We sample $N=1{,}000$ prompts $x_1,\dots,x_N$ from \texttt{tulu-3-sft-mixture}~\cite{lambert2025tulu} and use only the prompts for generation.

\paragraph{Estimation of KL divergence.}
The two KL estimates are computed in the same way as in Appendix~\ref{app:details-empirical-validation}. As in Section~\ref{sec:monte-carlo-kl}, we symmetrize the Monte Carlo estimate as $\frac{1}{2}\bigl(\mathrm{KL}(p_i,p_j)+\mathrm{KL}(p_j,p_i)\bigr)$ and compute Spearman's $\rho$ and Pearson's $r$ over all $276 = 24 \times 23 / 2$ unordered model pairs. For reference, we also compute the correlations without symmetrization over all $552 = 24 \times 23$ ordered model pairs, excluding the diagonal. In the latter comparison, the log-likelihood-vector-based estimate is identical for the two directions of each model pair, whereas the directional Monte Carlo estimates can differ.

\paragraph{Results.}
The symmetrized estimates are strongly correlated (Spearman's $\rho=0.834$; Pearson's $r=0.786$). Although lower than the $\rho=0.94$ observed with eight models in Section~\ref{sec:monte-carlo-kl}, the rank correlation remains substantial. Without symmetrization, the correlations decrease to $\rho=0.710$ and $r=0.659$, consistent with the directional mismatch described above.

\begin{table}[!t]
  \centering
  \begin{adjustbox}{width=\linewidth}
  \begin{tabular}{llrrr}
    \toprule
    KL & Model pairs & \#pairs & Spearman $\rho$ & Pearson $r$ \\
    \midrule
    Symmetrized & All                & 276 & 0.834 & 0.786 \\
                & Same family        &  58 & 0.908 & 0.909 \\
                & Different families & 218 & 0.781 & 0.732 \\
    \midrule
    {w/o Symmetrization} & All                & 552 & 0.710 & 0.659 \\
                & Same family        & 116 & 0.865 & 0.830 \\
                & Different families & 436 & 0.617 & 0.577 \\
    \bottomrule
  \end{tabular}
  \end{adjustbox}
  \caption{Correlation between the Monte Carlo estimate of KL divergence and the estimate based on log-likelihood vectors for the $K'=24$ models.}
  \label{tab:app-24models-corr}
\end{table}

\paragraph{Effect of model diversity.}
Table~\ref{tab:app-24models-corr} shows the correlations grouped by whether the two models in each pair belong to the same family. The $24$ models are grouped into six families according to their base models: Qwen ($9$ models), Llama ($6$), Gemma ($3$), Mistral ($3$), Phi ($2$), and Yi ($1$), which yields $58$ same-family and $218$ different-family unordered pairs. Under both conventions, the correlation is higher for same-family pairs (symmetrized: $\rho=0.908$, $r=0.909$) than for different-family pairs (symmetrized: $\rho=0.781$, $r=0.732$). This is consistent with the assumption underlying the variance approximation in Section~\ref{sec:kl-divergence}, namely that the models are sufficiently close to the data-generating distribution. It also suggests that the lower correlation compared to Section~\ref{sec:monte-carlo-kl} reflects the greater diversity of the present model set.

\section{Estimation of CO2 Emissions}
\label{app:co2}

We estimate the CO$_2$ emissions associated with computing the log-likelihood vectors using CarbonTracker v2.4.5~\citep{anthony2020carbontracker}.
We measure the GPU energy consumption for a representative sample of models on a single NVIDIA A6000 GPU; the measured cost is approximately 115~Wh per model per $10{,}000$ texts.
Scaling this to $552$ models and $5$ prompt settings gives approximately $317$~kWh in total.
Converting this energy using the average grid carbon intensity of Japan, where the experiments are conducted, yields approximately $153$~kgCO$_2$eq.\footnote{We use $482.32$~gCO$_2$eq/kWh, the 2024 country-average carbon intensity for Japan bundled with CarbonTracker.}


\begin{thebibliography}{20}
\providecommand{\natexlab}[1]{#1}

\bibitem[{Anthony et~al.(2020)Anthony, Kanding, and
  Selvan}]{anthony2020carbontracker}
Lasse F.~Wolff Anthony, Benjamin Kanding, and Raghavendra Selvan. 2020.
\newblock Carbontracker: Tracking and predicting the carbon footprint of
  training deep learning models.
\newblock ICML Workshop on Challenges in Deploying and monitoring Machine
  Learning Systems.
\newblock ArXiv:2007.03051.

\bibitem[{Ding et~al.(2023)Ding, Chen, Xu, Qin, Hu, Liu, Sun, and
  Zhou}]{ding-etal-2023-enhancing}
Ning Ding, Yulin Chen, Bokai Xu, Yujia Qin, Shengding Hu, Zhiyuan Liu, Maosong
  Sun, and Bowen Zhou. 2023.
\newblock \href {https://doi.org/10.18653/v1/2023.emnlp-main.183} {Enhancing
  chat language models by scaling high-quality instructional conversations}.
\newblock In \emph{Proceedings of the 2023 Conference on Empirical Methods in
  Natural Language Processing}, pages 3029--3051, Singapore. Association for
  Computational Linguistics.

\bibitem[{Horwitz et~al.(2025)Horwitz, Kurer, Kahana, Amar, and
  Hoshen}]{horwitz2025we}
Eliahu Horwitz, Nitzan Kurer, Jonathan Kahana, Liel Amar, and Yedid Hoshen.
  2025.
\newblock \href {https://openreview.net/forum?id=BzFMBNqg7R} {{We Should Chart
  an Atlas of All the World's Models}}.
\newblock In \emph{The Thirty-Ninth Annual Conference on Neural Information
  Processing Systems Position Paper Track}.

\bibitem[{Jiang et~al.(2025)Jiang, Chai, Li, Liu, Fok, Dziri, Tsvetkov, Sap,
  and Choi}]{jiang2025artificial}
Liwei Jiang, Yuanjun Chai, Margaret Li, Mickel Liu, Raymond Fok, Nouha Dziri,
  Yulia Tsvetkov, Maarten Sap, and Yejin Choi. 2025.
\newblock \href {https://openreview.net/forum?id=saDOrrnNTz} {{Artificial
  Hivemind: The Open-Ended Homogeneity of Language Models (and Beyond)}}.
\newblock In \emph{The Thirty-ninth Annual Conference on Neural Information
  Processing Systems Datasets and Benchmarks Track}.

\bibitem[{Kishino et~al.(2026)Kishino, Takase, Oyama, Yamagiwa, and
  Shimodaira}]{kishino-etal-2026-establishing}
Ryo Kishino, Yusuke Takase, Momose Oyama, Hiroaki Yamagiwa, and Hidetoshi
  Shimodaira. 2026.
\newblock \href {https://doi.org/10.18653/v1/2026.findings-acl.1163}
  {Establishing a scale for {K}ullback-{L}eibler divergence in language models
  across various settings}.
\newblock In \emph{Findings of the {A}ssociation for {C}omputational
  {L}inguistics: {ACL} 2026}, pages 23223--23248, San Diego, California, United
  States. Association for Computational Linguistics.

\bibitem[{Kojima et~al.(2022)Kojima, Gu, Reid, Matsuo, and
  Iwasawa}]{kojima2022large}
Takeshi Kojima, Shixiang~Shane Gu, Machel Reid, Yutaka Matsuo, and Yusuke
  Iwasawa. 2022.
\newblock Large language models are zero-shot reasoners.
\newblock \emph{Advances in neural information processing systems},
  35:22199--22213.

\bibitem[{Kwon et~al.(2023)Kwon, Li, Zhuang, Sheng, Zheng, Yu, Gonzalez, Zhang,
  and Stoica}]{kwon2023efficient}
Woosuk Kwon, Zhuohan Li, Siyuan Zhuang, Ying Sheng, Lianmin Zheng, Cody~Hao Yu,
  Joseph Gonzalez, Hao Zhang, and Ion Stoica. 2023.
\newblock \href {https://doi.org/10.1145/3600006.3613165} {Efficient memory
  management for large language model serving with pagedattention}.
\newblock In \emph{Proceedings of the 29th Symposium on Operating Systems
  Principles}, SOSP '23, page 611–626, New York, NY, USA. Association for
  Computing Machinery.

\bibitem[{Lambert et~al.(2025)Lambert, Morrison, Pyatkin, Huang, Ivison,
  Brahman, Miranda, Liu, Dziri, Lyu, Gu, Malik, Graf, Hwang, Yang, Bras,
  Tafjord, Wilhelm, Soldaini, Smith, Wang, Dasigi, and
  Hajishirzi}]{lambert2025tulu}
Nathan Lambert, Jacob Morrison, Valentina Pyatkin, Shengyi Huang, Hamish
  Ivison, Faeze Brahman, Lester James~Validad Miranda, Alisa Liu, Nouha Dziri,
  Xinxi Lyu, Yuling Gu, Saumya Malik, Victoria Graf, Jena~D. Hwang, Jiangjiang
  Yang, Ronan~Le Bras, Oyvind Tafjord, Christopher Wilhelm, Luca Soldaini, and
  4 others. 2025.
\newblock \href {https://openreview.net/forum?id=i1uGbfHHpH} {Tulu 3: Pushing
  frontiers in open language model post-training}.
\newblock In \emph{Second Conference on Language Modeling}.

\bibitem[{Leviathan et~al.(2025)Leviathan, Kalman, and
  Matias}]{leviathan2025prompt}
Yaniv Leviathan, Matan Kalman, and Yossi Matias. 2025.
\newblock Prompt repetition improves non-reasoning {LLM}s.
\newblock \emph{arXiv preprint arXiv:2512.14982}.

\bibitem[{Li et~al.(2025)Li, Du, Zhao, Zhang, Wang, Gao, Liu, and
  Lin}]{li2025infinity}
Jijie Li, Li~Du, Hanyu Zhao, {Bo-wen} Zhang, Liangdong Wang, Boyan Gao, Guang
  Liu, and Yonghua Lin. 2025.
\newblock \href {https://arxiv.org/abs/2506.11116} {{I}nfinity {I}nstruct:
  Scaling instruction selection and synthesis to enhance language models}.
\newblock \emph{Preprint}, arXiv:2506.11116.

\bibitem[{Mikolov et~al.(2013)Mikolov, Sutskever, Chen, Corrado, and
  Dean}]{mikolov2013distributed}
Tomas Mikolov, Ilya Sutskever, Kai Chen, Greg~S Corrado, and Jeff Dean. 2013.
\newblock \href
  {https://proceedings.neurips.cc/paper_files/paper/2013/file/9aa42b31882ec039965f3c4923ce901b-Paper.pdf}
  {Distributed representations of words and phrases and their
  compositionality}.
\newblock In \emph{Advances in Neural Information Processing Systems},
  volume~26. Curran Associates, Inc.

\bibitem[{Oyama et~al.(2025{\natexlab{a}})Oyama, Kishino, Yamagiwa, and
  Shimodaira}]{oyama-etal-2025-likelihood}
Momose Oyama, Ryo Kishino, Hiroaki Yamagiwa, and Hidetoshi Shimodaira.
  2025{\natexlab{a}}.
\newblock \href {https://doi.org/10.18653/v1/2025.findings-emnlp.502}
  {{Likelihood Variance as Text Importance for Resampling Texts to Map Language
  Models}}.
\newblock In \emph{Findings of the Association for Computational Linguistics:
  EMNLP 2025}, pages 9453--9465, Suzhou, China. Association for Computational
  Linguistics.

\bibitem[{Oyama et~al.(2025{\natexlab{b}})Oyama, Yamagiwa, Takase, and
  Shimodaira}]{modelmap2025}
Momose Oyama, Hiroaki Yamagiwa, Yusuke Takase, and Hidetoshi Shimodaira.
  2025{\natexlab{b}}.
\newblock \href {https://aclanthology.org/2025.acl-long.1584/} {{Mapping 1,000+
  Language Models via the Log-Likelihood Vector}}.
\newblock In \emph{Proceedings of the 63rd Annual Meeting of the Association
  for Computational Linguistics (Volume 1: Long Papers)}, pages 32983--33038.

\bibitem[{Reimers and Gurevych(2019)}]{reimers-gurevych-2019-sentence}
Nils Reimers and Iryna Gurevych. 2019.
\newblock \href {https://doi.org/10.18653/v1/D19-1410} {Sentence-{BERT}:
  Sentence embeddings using {S}iamese {BERT}-networks}.
\newblock In \emph{Proceedings of the 2019 Conference on Empirical Methods in
  Natural Language Processing and the 9th International Joint Conference on
  Natural Language Processing (EMNLP-IJCNLP)}, pages 3982--3992, Hong Kong,
  China. Association for Computational Linguistics.

\bibitem[{Shimodaira(1993)}]{Shimodaira1993modelmap}
Hidetoshi Shimodaira. 1993.
\newblock \href {https://ismrepo.ism.ac.jp/records/31997} {A model search
  technique based on confidence set and map of models
  [モデルの信頼集合と地図によるモデル探索] (in {Japanese})}.
\newblock \emph{Proceedings of the Institute of Statistical Mathematics},
  41(2):131--147.

\bibitem[{Shimodaira and Cao(1998)}]{shimodaira1998graphical}
Hidetoshi Shimodaira and Ying Cao. 1998.
\newblock A graphical technique for model selection diagnosis.
\newblock \emph{The Institute of Statistical Mathematics Research Memorandum},
  680.

\bibitem[{Yax et~al.(2025)Yax, Oudeyer, and Palminteri}]{yax2025phylolm}
Nicolas Yax, Pierre-Yves Oudeyer, and Stefano Palminteri. 2025.
\newblock \href {https://openreview.net/forum?id=rTQNGQxm4K} {Phylo{LM}:
  Inferring the phylogeny of large language models and predicting their
  performances in benchmarks}.
\newblock In \emph{Proceedings of the 13th International Conference on Learning
  Representations}.

\bibitem[{Zhou et~al.(2025)Zhou, Chen, Cahyawijaya, Duan, and
  Cai}]{zhou-etal-2025-linguistic}
Xinyu Zhou, Delong Chen, Samuel Cahyawijaya, Xufeng Duan, and Zhenguang Cai.
  2025.
\newblock \href {https://aclanthology.org/2025.coling-main.459/} {Linguistic
  minimal pairs elicit linguistic similarity in large language models}.
\newblock In \emph{Proceedings of the 31st International Conference on
  Computational Linguistics}, pages 6866--6888, Abu Dhabi, UAE. Association for
  Computational Linguistics.

\bibitem[{Zhu et~al.(2025)Zhu, Ahmed, Kuditipudi, and
  Liang}]{zhu2025independence}
Sally Zhu, Ahmed~M Ahmed, Rohith Kuditipudi, and Percy Liang. 2025.
\newblock \href {https://openreview.net/forum?id=mzSwYvwYdC} {Independence
  tests for language models}.
\newblock In \emph{Proceedings of the 42nd International Conference on Machine
  Learning}.

\bibitem[{Zhuang et~al.(2025)Zhuang, Wu, Wen, Li, Jiao, and
  Ramchandran}]{zhuang2025embedllm}
Richard Zhuang, Tianhao Wu, Zhaojin Wen, Andrew Li, Jiantao Jiao, and Kannan
  Ramchandran. 2025.
\newblock \href {https://openreview.net/forum?id=Fs9EabmQrJ} {Embed{LLM}:
  Learning compact representations of large language models}.
\newblock In \emph{Proceedings of the 13th International Conference on Learning
  Representations}.

\end{thebibliography}
\end{document}